\documentclass[conference]{IEEEtran}
\IEEEoverridecommandlockouts
\usepackage{cite}
\usepackage{algpseudocode} 
\usepackage{balance}
\usepackage{soul}
\usepackage{array}
\usepackage[nointegrals]{wasysym}
\usepackage{tikz}
\usepackage{times}
\usepackage{soul}
\usepackage{url}
\usepackage[hidelinks]{hyperref}
\usepackage[utf8]{inputenc}
\usepackage[small]{caption}
\usepackage{graphicx}
\usepackage{booktabs}
\usepackage{dsfont}
\usepackage{amssymb}
\usepackage{amsmath}
\usepackage{multirow}
\usepackage{enumitem}
\usepackage{upgreek}
\usepackage[ruled,linesnumbered]{algorithm2e}
\usepackage{xcolor}
\usepackage{soul}
\usepackage[font=small,skip=5pt]{caption}
\usepackage[font=small,skip=5pt,labelformat=simple]{subcaption}
\usepackage{float}
\usepackage{diagbox}
\usepackage{pbox}
\usepackage{dblfloatfix}

\usepackage{amsmath}
\usepackage{hhline}

\def\BibTeX{{\rm B\kern-.05em{\sc i\kern-.025em b}\kern-.08em
    T\kern-.1667em\lower.7ex\hbox{E}\kern-.125emX}}

\def\revision#1{{#1}}

\begin{document}

\title{Towards Spatio-Temporal Aware Traffic Time Series Forecasting--Full Version}

\author{
    \IEEEauthorblockN{Razvan-Gabriel Cirstea$^1$, Bin Yang$^{1}$
    Chenjuan Guo$^1$, Tung Kieu$^1$, Shirui Pan$^{2}$ }
    \IEEEauthorblockA{$^1$Department of Computer Science, Aalborg University, Denmark \\
    $^2$Faculty of Information Technology, Monash University, Australia\\
    \{razvan, byang, cguo, tungkvt \}@cs.aau.dk, shirui.pan@monash.edu
    }
}

\maketitle
\thispagestyle{plain}
\pagestyle{plain}

\begin{abstract}
Traffic time series forecasting is challenging due to complex \textit{spatio-temporal dynamics}---time series from different locations often have distinct patterns; and for the same time series, patterns may vary across time, where, for example, there exist certain periods across a day showing stronger temporal correlations. 
Although recent forecasting models, in particular deep learning based models, show promising results, they 
suffer from being \emph{spatio-temporal agnostic}. 
Such spatio-temporal agnostic models 
employ a shared parameter space irrespective of the time series locations and the time periods 
and they assume that the temporal patterns are similar across locations and do not evolve across time, which may not always hold, 
thus leading to sub-optimal results. In this work, we propose a framework that aims at turning spatio-temporal agnostic models to \emph{spatio-temporal aware} models. To do so, we encode time series from different locations into stochastic variables, from which we generate location-specific and time-varying model parameters to better capture the spatio-temporal dynamics. 
We show how to integrate the framework with canonical attentions to enable spatio-temporal aware attentions. 
Next, to compensate for the additional overhead introduced by the spatio-temporal aware model parameter generation process, we propose a novel window attention scheme,  which helps reduce the complexity from quadratic to linear, making spatio-temporal aware attentions also have competitive efficiency. 
We show strong empirical evidence on four traffic time series datasets, where the proposed spatio-temporal aware attentions outperform state-of-the-art methods in term of accuracy and efficiency.
This is an extended version of ``Towards Spatio-Temporal Aware Traffic Time Series Forecasting'', to appear in ICDE 2022~\cite{Razvanicde2022}, including additional experimental results.

\end{abstract}

\section{Introduction}

\begin{figure}[t]
    \centering
        \begin{subfigure}[t]{.95\linewidth}
        \centering
            \includegraphics[width=.85\columnwidth,height=3cm]{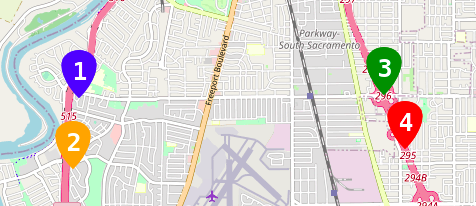}
            \caption{Sensors Locations, Sacramento, California.
            }
            \label{fig:sensor_locations}
        \end{subfigure}
        \\
        \begin{subfigure}[t]{.99\linewidth}
        \centering
            \includegraphics[width=\columnwidth]{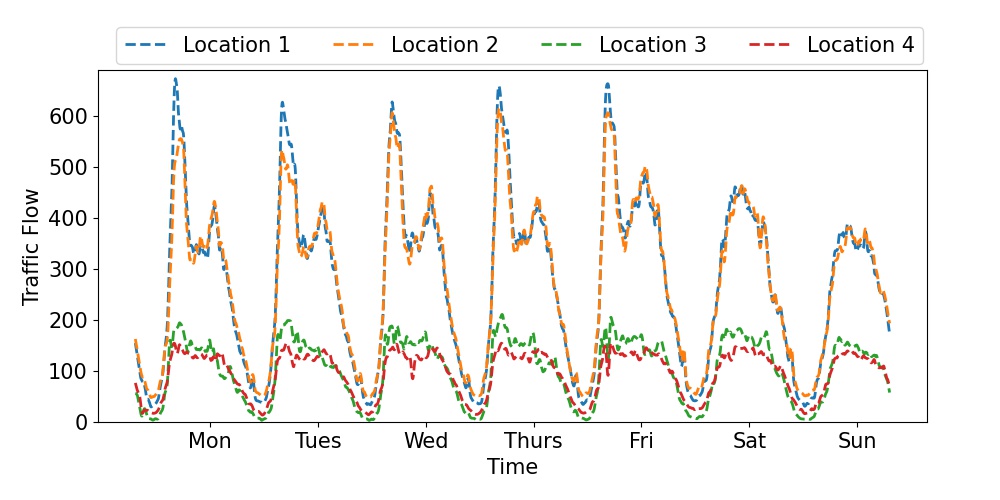}
            \vspace{-5pt}
            \caption{Time series from the above 4 sensors, showing 1-week traffic conditions}
            \label{fig:traffic}
        \end{subfigure}
        \caption{Example of four sensors and their time series. The time series from sensors 1 and 2 show similar traffic patterns as the sensors are deployed on the same street. The same happens for sensors 3 and 4, but the time series from sensors 1 and 3 show different patterns.
    }
    \label{fig:traffic_example}
\end{figure}

We are witnessing impressive technological developments in the last years, leading to inexpensive and effective monitoring systems using a wide variety of sensors~\cite{tkdesean,DBLP:conf/cikm/Kieu0GJ18}. For example, the transportation sector has deployed different sensors, e.g., loop detectors or speed cameras, in different roads to continuously capture useful traffic information, such as speeds and traffic flows, 
at different roads, giving rise to a large amount of traffic time series. 
%
An example is shown 
in Figure~\ref{fig:traffic_example}, where we plot the physical locations of four traffic flow sensors deployed on the highways of Sacramento, California, and the traffic flow time series collected by these sensors. 

Traffic forecasting using traffic time series collected by the sensors is a core component of many Intelligent Transportation Systems (ITS)~\cite{DBLP:conf/ijcai/YangGHT021}. 
%
Accurate forecasting on traffic statuses, e.g., traffic flow, speeds, lane occupancy, reveals holistic dynamics of the underlying traffic network, contributing to early warnings for emergency management~\cite{zheng2014urban,DBLP:conf/mdm/Kieu0J18}, diagnosing faulty equipment~\cite{tungicde2022, davidpvldb,tungicde2022second}, and route planning and recommendation~\cite{DBLP:journals/vldb/PedersenYJ20,DBLP:journals/vldb/GuoYHJC20,DBLP:conf/icde/LiuJYZ18,DBLP:journals/pvldb/PedersenYJ20,SeanIcde2022}. 
However, accurate traffic forecasting is challenging due to its complex \emph{spatio-temporal dynamics}.

First, time series collected from different locations have distinct patterns. 
To illustrate this, 
we show four different time series collected from the four sensors shown in Figure~\ref{fig:sensor_locations}.  
%
Sensors 1 (blue) and 2 (orange), which are deployed along one street, exhibit different patterns, comparing to  sensors 3 (green) and 4 (red), which are deployed along another street.  
%
More specifically, sensors 1 and 2 have two clear peek hours in the morning and in the afternoon, respectively, in weekdays; while sensors 3 and 4 do not have a spike in the afternoon, instead the flow gradually decreases. 
%
%
%
This calls for \emph{spatial-aware modeling} where different model parameters can be used for modeling time series from different locations.  

Second, the patterns within time series often vary across time. 
Figure~\ref{fig:sensor_locations} shows that during weekdays, it exhibits similar temporal patterns, which are different from the patterns during the weekends. This holds for both sets of sensors. In such scenarios, it can be beneficial to automatically adjust model parameters across time such that they can better model different temporal patterns in, e.g., weekdays vs. weekends. 
The ability of adjusting model parameters across time can also be useful for %
accidents or road closures, where traffic patterns may deviate from regular temporal patterns. 
This calls for \emph{temporal-aware modeling} where different model parameters can be used for modeling time series during different times. 

Based on the above, the complex spatio-temporal traffic dynamics calls for spatio-temporal aware forecasting model that is able to (1) capture distinct patterns at different locations; (2) quickly adapt to pattern changes in different time periods. 


However, existing methods are often \emph{spatio-temporal agnostic}, thus failing to capture the spatio-temporal dynamics. Traditional methods such as Auto-Regressive Integrated Moving Average (\texttt{ARIMA}) or Vector Auto-Regression (\texttt{VAR}) cannot capture nonlinear patterns and thus fail to capture complex spatio-temporal patterns among different time series, resulting in sub-optimal forecasting accuracy~\cite{arima_non_linear}. Recently, various deep learning models have been proposed with promising accuracy, which are mainly based on three types of architectures, i.e., recurrent neural networks (\texttt{RNN}s)~\cite{bai2019passenger,dcrnn,agcrn,lstm_gcn,DBLP:conf/icde/Hu0GJX20}, temporal convolution networks (\texttt{TCN}s)~\cite{tcn,graphwavenet,stgcn}, and attentions~\cite{gman,st-grat,li2019enhancing}. However, none of the architectures offers time-varying and location-specific modeling. 
First, the model parameters are static, thus failing to capture patterns that change over time and being \emph{temporal-agnostic}. In particular, the weight matrices in \texttt{RNN} cells, the convolution filters in \texttt{TCN}s, and the projection matrices in attentions, stay the same across time~\cite{bai2019passenger,dcrnn,graphwavenet,dual_graph_convolutional,stfgnn,astgcn}. 
Second, the same parameters are employed for time series from different locations~\cite{dcrnn,graphwavenet,stsgcn,bai2019passenger,astgcn}, by assuming that the traffic patterns are similar for all locations. However, this is often not the case in reality as shown in Figure~\ref{fig:traffic_example}. Such modeling forces the learned parameters to represent an ``average'' of the traffic patterns among all time series, thus being \emph{spatial-agnostic}.

\revision{In this paper, we strive to enable \textit{spatio-temporal aware} forecasting models by utilizing time-varying and location-specific model parameters in a data driven manner. }
We propose to encode each location’s recent time series into a stochastic variable that captures both location-specific and time-varying patterns, from which we generate distinct model parameters for modeling time series from different locations and for different time periods. \revision{This approach is purely data driven, which avoids using additional location and time specific features, such as coordinates of sensor locations and prior knowledge on peak vs. offpeak hours. }
This approach is generic as it can be applied to generate spatio-temporal aware parameters for different models, such as \texttt{RNNs}, \texttt{TCN}s and attentions. We use attention as a concrete model in this paper as it shows superior accuracy, even for very long time series~\cite{li2019enhancing}, compared to, e.g., \texttt{RNNs} and \texttt{TCNs}. 


To alleviate the additional overhead caused by the spatio-temporal aware model parameter generation, we propose a novel \revision{efficient and accurate} attention mechanism to ensure that the spatio-temporal aware attentions also have competitive efficiency. 
More specifically, in canonical self-attention, each timestamp directly attends all other timestamps, thus resulting in quadratic 
complexity. 
%
%
We propose a \textit{window attention} to reduce the complexity from quadratic to linear, thus ensuring competitive efficiency overall, \revision{without compromising accuracy. }
More specifically, we strategically break an input time series into small windows and introduce a small constant number of \emph{proxies} for each window to harvest information within the window. 
Then, each timestamp only attends to the proxies, but not all other timestamps anymore.  
This enables us to reduce the complexity from  $\mathcal{O}(H^2)$ of the canonical self-attention to $\mathcal{O}(H)$ of the window attention, where $H$ is the length of the input time series. This linear window attention operation ensures competitive overall efficiency, even with the additional spatio-temporal aware parameter generation. 

To the best of our knowledge, this is the first study to enable spatio-temporal aware modeling where time-varying and location-specific model parameters are employed. 
The study makes four contributions. First, we propose Spatio-Temporal Aware Parameter Generation Network to generate \emph{location-specific} and \emph{time-varying} model parameters to turn spatio-temporal agnostic models to spatio-temporal aware models. 
Second, we integrate the proposed mechanism into attentions to enable spatio-temporal aware attentions. 
Third, we propose an efficient window attention by using proxies, reducing the complexity of attentions operations from quadratic to linear, thus ensuring overall competitive efficiency for spatio-temporal aware attentions. 
Fourth, we conduct extensive experiments on four commonly used traffic time series data sets, justifying the design choices and demonstrating that the proposal outperforms 
the state-of-the-art. 

\noindent
\emph{Paper Outline}: Section \ref{section:related_work} covers the recent advancements in the field. Section \ref{section:problem} formally defines the problem and presents the overview of the model. Section \ref{section:results} describe the experimental study and the results and \ref{section:conclusion} concludes.

\section{Related Work}\label{section:related_work}

\textbf{Time Series Forecasting.} 
To achieve accurate forecasting results, modeling temporal dynamics is an essential component of any time series forecasting model. Traditional methods such as Vector Auto-Regression (\texttt{VAR}) or Auto-Regressive Integrated Moving Average (\texttt{ARIMA}) \cite{arima} cannot capture nonlinear patterns and thus fail to capture complex temporal patterns among different time series, resulting in sub-optimal forecasting accuracy \cite{arima_non_linear}. Recent studies show that deep learning methods are able to consistently outperform traditional methods in the context of time series forecasting \cite{dcrnn}. \revision{Existing studies on deep learning time series forecasting often use three types of models---recurrent neural networks (\texttt{RNN}s)~\cite{bai2019passenger,dcrnn,agcrn,lstm_gcn,DBLP:conf/icde/Hu0GJX20,MileTS} in the form of \texttt{LSTM} or \texttt{GRU}, and temporal convolutions networks (\texttt{TCNs})~\cite{tcn,graphwavenet,stgcn}.}

Recently attention mechanism such as \texttt{Transformers}~\cite{gman,st-grat,li2019enhancing,astgnn,wupvldb} have shown superior performance when compared with \texttt{RNN} and \texttt{TCN} based models, as they are better at handling long term dependencies.
However, they suffer from quadratic memory and runtime overhead w.r.t. to the input sequence length $H$. To address those limitations, in sliding window attentions, e.g., \cite{beltagy2020longformer,zhang2021multi}, each timestamp considers a sliding window of length $S$, covering its past and future neighboring timestamps. As such, each timestamp attends only to the $S$ timestamps in the sliding window. This leads to $\mathcal{O}(H \times S)$ complexity, where $H$ is the length of time series. We instead strive for attentions with linear complexity $\mathcal{O}(H)$. Furthermore, existing attention models, including the sliding window attentions, are still \textit{spatial-temporal agnostic} as they use the same projection matrices irrespective of the time series locations and the time periods.

\textbf{Categorization of Deep Learning Based Forecasting.} 
We systematically review deep learning studies on time series forecasting from two aspects---spatial awareness vs. temporal awareness.
First, spatial agnostic models use the same set of model parameters for time series from different locations, whereas spatial aware models use distinct sets of model parameters for time series from different locations. 
Second, temporal agnostic models use the same set of model parameters for different time periods, whereas temporal aware models use distinct sets of model parameters for different time periods. 
%
%
The categorization is shown in Table \ref{table:related_work}. 

\begin{table}[h]
\centering
\refstepcounter{table}
\begin{tabular}{|c|c|c|} 
\toprule
\backslashbox{Spatial}{Temporal} & Agnostic & Aware  \\ 
\hline
Agnostic                    &    \pbox{3cm}{\cite{bai2019passenger,xu2020spatial,geng2019spatiotemporal,3dcnn,wang2020traffic,st-grat,gman,zhang2021multi} \textbf{\cite{dcrnn,astgcn,bai2019stg2seq,graphwavenet,stsgcn,stfgnn}},\revision{\textbf{\cite{beltagy2020longformer,astgnn}}}}      & \pbox{2cm} {\textbf{\cite{chen2018meta}}}         \\ 
\hline
Aware          &  \cite{KDD_urban_traffic,DBLP:journals/pvldb/0002GJ13},\textbf{\cite{enhancenet,agcrn}}       &     \textbf{Ours}     \\
\bottomrule
\end{tabular}
\caption{Categorization of Related Studies. The bold references are included in the empirical study, covering the strongest and latest baselines in each category.}
\label{table:related_work}
\end{table}




Most of the related studies are \emph{spatio-temporal agnostic}, as shown in the top-left corner of Table \ref{table:related_work}.  They apply the same model parameters, e.g., projection matrices in self-attention, weight matrices in \texttt{RNNs}, convolution filters in \texttt{TCNs}, for all time series from different locations. By using the same parameter space for all locations such studies implicitly assume that the traffic patterns are similar among locations. Such assumption may not always hold in practice, as demonstrated by a real world example in Figure \ref{fig:traffic_example}, and this inhibits the model capabilities as the models parameters only learn to capture the ``average'' traffic dynamics among all locations.

Three existing studies~\cite{KDD_urban_traffic,agcrn,enhancenet} propose \textit{spatial-aware} but still \textit{temporal-agnostic}, \texttt{RNN} based models by using location-specific weight matrices for \texttt{RNNs}.  Specifically, \cite{KDD_urban_traffic} generates location-specific weight matrices for \texttt{RNNs} based on additional meta-knowledge, e.g., categories of points of interest around the deployed sensor locations which might not always be available, limiting the applicability of the model, e.g., in the data sets where such additional meta-knowledge is unavailable. \cite{agcrn} learns a pool of candidate weights and choose to combine them differently to obtain location-specific weight matrices for \texttt{RNNs}. 
One recent 
study~\cite{enhancenet} employs a deterministic memory per location to generate spatial-aware model parameters 
for time series forecasting, but still fails to enable temporal-aware modeling. In contrast, our proposal proposes to employ a stochastic variable to generate both spatial and temporal aware model parameters. 
Thus, \cite{enhancenet} can be considered as a special case of the paper with the standard deviation of the stochastic variable being always 0 and without temporal-aware parameter generation. 
We show strong empirical results that our proposal outperforms \cite{agcrn} and \cite{enhancenet} (see Table \ref{table:results}).  Spatial-aware model parameter generation has also been studied for image recognition and object detection  ~\cite{chen2020dynamic,dynamic_filters_network,yang2019condconv}, but fail to capture temporal dynamics and thus cannot be applied directly to enable time series forecasting.

Finally, one existing study \cite{chen2018meta} is \textit{temporal-aware} but still \textit{spatial-agnostic}. The authors propose to use an additional \texttt{LSTM} which acts as a $meta$-\texttt{LSTM}. The $meta$-\texttt{LSTM}'s hidden representation, which varies across time, is then used to generate time-varying model parameters for another \texttt{LSTM}. 
The empirical study shows that our proposal outperforms \cite{chen2018meta}. 


\section{Preliminaries}

\begin{table}
\centering

\begin{tabular}{|c|l|} 
\hline
Variable                             & Definition                                               \\ 
\hhline{|==|}
\textbf{X}                           & Time series data                                         \\ 
\hline
$H$                                  & Length of the input time series                          \\ 
\hline
$U$                                  & Length of the predicted time series                      \\ 
\hline
$N$                                  & Number of sensors/time series                            \\ 
\hline
$\textbf{h}^(i)$                     & Output of attention for the \textit{i}-th sensor         \\ 
\hline
$\mathbf{\Theta}_t^(i)$              & Stochastic latent variable for the \textit{i}-th sensor  \\ 
\hline
$\textbf{z}^(i)$                     & Spatial-aware stochastic variable                        \\ 
\hline
$\textbf{z}^(i)_t$                   & Temporal-aware stochastic variable                       \\ 
\hline
$E_\psi$                             & An encoder network $E$ parameterized by~$\psi$           \\ 
\hline
$D_\omega$                           & A decoder network $D$ ~parameterized by $\omega$         \\ 
\hline
$W$                                  & Number of windows                                        \\ 
\hline
$S$                                  & Window size~                                             \\ 
\hline
$\textbf{P}$                         & Proxy tensor                                             \\ 
\hline
$\textbf{Q}, \textbf{K}, \textbf{V}$ & Learnable projection matrices in attentions              \\
\hline
\end{tabular}
\caption{Table of notations}
\vspace{-2em}
\label{table:notations}
\end{table}

\subsection{ Problem Definition}\label{section:problem}
Given a set of $N$ sensors that are deployed in different locations in a spatial network, each sensor produces a multidimensional time series recording $F$ attributes (e.g., traffic speed, traffic flow) across time. Assuming that each time series has a total of $T$ timestamps, we use $\textbf{X} \in \mathbb{R}^{N\times T \times F}$ to represent the time series from all $N$ sensors. We use $\mathbf{x}^{(i)} \in \mathbb{R}^{T\times F}$ to denote the  time series from the $i$-th sensor, $\mathbf{x}_{t} \in \mathbb{R}^{N\times F}$ to denote the attributes from all sensors at the $t$-th timestamp, and $\mathbf{x}^{(i)}_t \in \mathbb{R}^{F}$ to represent the attribute vector from the $i$-th sensor at the $t$-th timestamp. 

Time series forecasting learns a function $\mathcal{F}$ that, at timestamp $t$, given the attributes of the past $H$ timestamps from all sensors, predicts the attributes of all sensors in the future $U$ timestamps.

\vspace{-5pt}
\begin{equation}
\mathcal{F}_\phi (\mathbf{x}_{t-H+1},..., \mathbf{x}_{t-1},\mathbf{x}_{t}) = (\mathbf{\hat{x}}_{t+1}, \mathbf{\hat{x}}_{t+2},...,\mathbf{\hat{x}}_{t+U} ),
\end{equation}
where $\phi$ denotes the learnable model parameters of the forecasting model and $\mathbf{\hat{x}}_{j}$ is the prediction at time $j$. \revision{Important notation is shown in Table \ref{table:notations}.}

\subsection{Spatio-temporal Agnostic Attention based Forecasting}

\begin{figure}[H]
    \centering
    \includegraphics[width=0.9\linewidth]{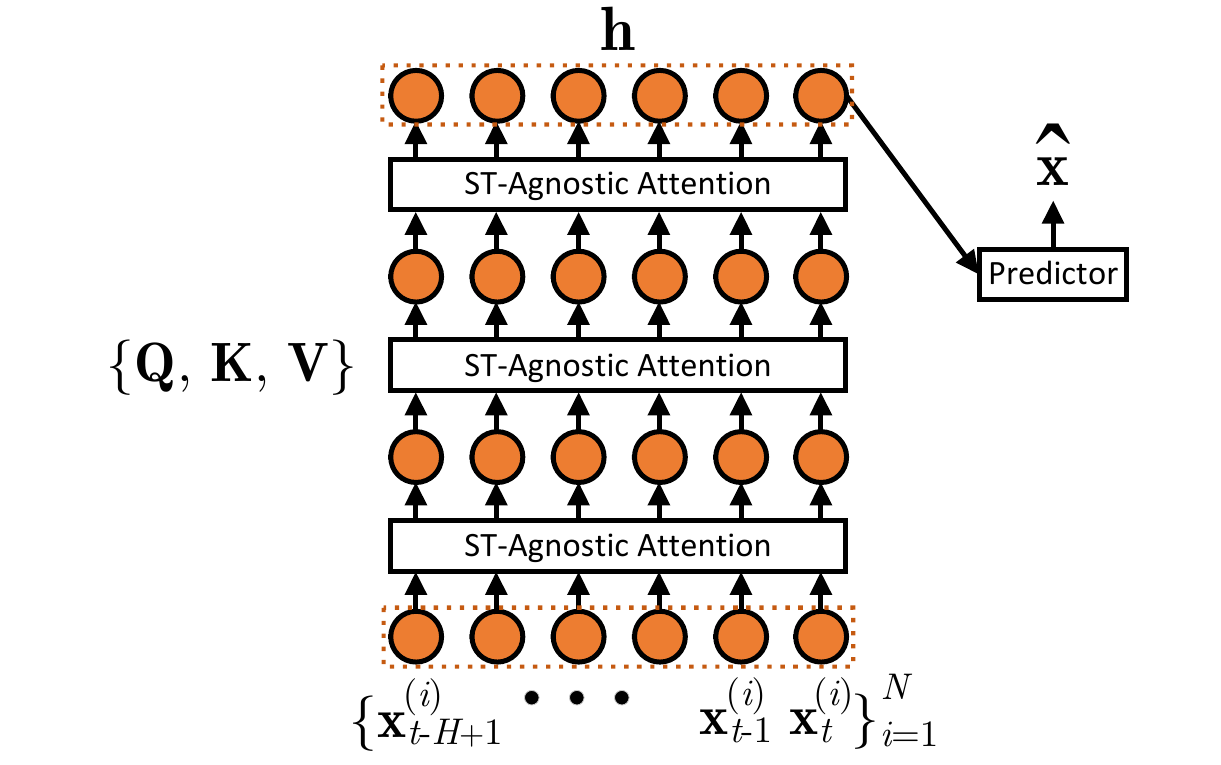}
    \caption{Spatio-temporal Agnostic Forecasting Model.}
    \label{fig:stagnostic}
\end{figure}

We use canonical attention~\cite{attention_all_you_need}, more specifically, self-attention, which is \emph{spatio-temporal agnostic}, as an example to illustrate the needs of \emph{spatio-temporal aware} modeling for traffic forecasting. 
%
Figure~\ref{fig:stagnostic} shows an overview of self-attention based forecasting, where multiple attention layers are stacked and then a predictor, e.g., a neural network, is applied on the output of the last layer to make forecasts $\mathbf{\hat{X}}$. 

In the following, we first cover how to use self-attention to make forecasting for a single time series and multiple time series, respectively, and then illustrate the needs of spatio-temporal aware modeling. 

\begin{figure*}[t]
    \centering
    \includegraphics[width=0.9\textwidth]{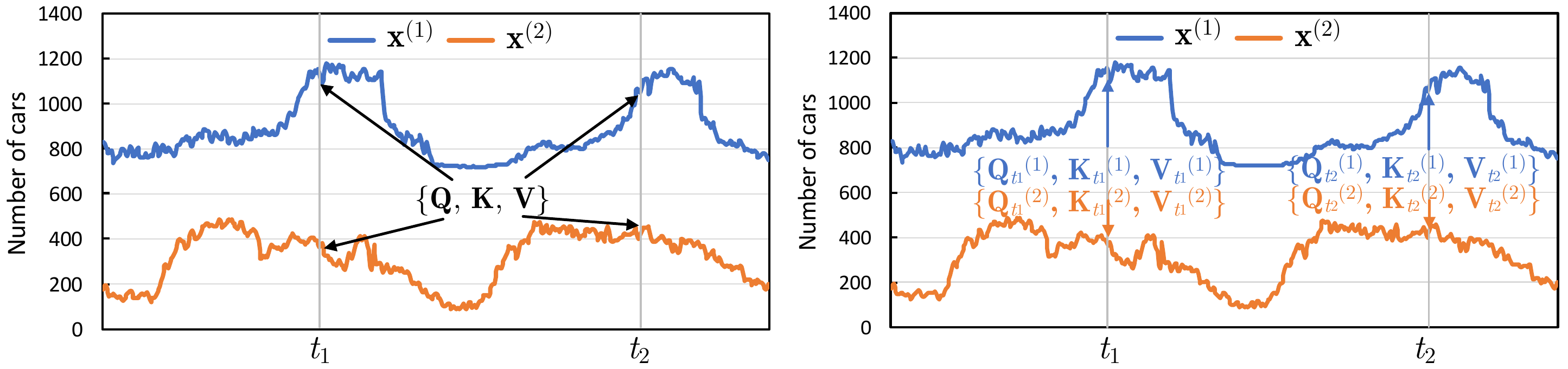}
    \caption{Spatio-temporal Agnostic Attentions (Left) use the \emph{same} set of projection matrices for different time series across different times. Spatio-temporal Aware Attentions (Right) use \emph{distinct} sets of projection matrices for different time series across different times.}
    \label{fig:stawareaaa}
    \vspace{-8pt}
\end{figure*}

\noindent
\textit{Single Time Series}: Consider an input time series with $H$ timestamps, e.g., time series $\mathbf{x}^{(i)}\in\mathbf{R}^{H\times F}$ from the $i$-th sensor. Self-attention first transforms the time series into a query matrix $\mathbf{x}^{(i)}\mathbf{Q}$, a key matrix $\mathbf{x}^{(i)}\mathbf{K}$ and a value matrix $\mathbf{x}^{(i)}\mathbf{V}$. Here,  \textit{projection matrices}
%
$\mathbf{Q}, \mathbf{K}, \mathbf{V} \in \mathbb{R}^{F \times d}$ 
are the learnable model parameters of self-attentions.


The output $\textbf{h}^{(i)}\in\mathbb{R}^{H \times d}$ of self-attention is represented as a  weighted sum of the values in the value matrix, where the weights, a.k.a., attention scores,  represent pairwise similarities between any two timestamps and are computed based on the query matrix and the key matrix, as shown in Equation~\ref{eq:attention}. 

\begin{equation} \label{eq:attention}
\begin{split}
  \textbf{h}^{(i)} &= \mathcal{A}tt (\textbf{x}^{(i)} \; | \; \textbf{Q}, \textbf{K}, \textbf{V}) 
  \\ &= 
  \sigma  (\frac{ ( \mathbf{x}^{(i)}  \mathbf{Q})   ( \mathbf{x}^{(i)} \textbf{K})^T}{\sqrt{d}} ) (\mathbf{x}^{(i)} \mathbf{V}),
\end{split} 
 \end{equation}
where $\sigma$ represents the softmax activation function. 

\noindent
\textit{Multiple Time Series}: We now consider the time series from all $N$ censors, upon which we can rewrite self-attentions as shown in Equation \ref{eq:attention_multiple}.     
\begin{equation} \label{eq:attention_multiple}
 \begin{split}
  \textbf{h} &= \mathcal{A}tt (\textbf{x} \; | \; \mathbf{Q}, \mathbf{K}, \mathbf{V}) 
  \\ &=  \left\{ 
 \sigma  (\frac{ ( \mathbf{x}^{(i)}  \mathbf{Q})   ( \mathbf{x}^{(i)} \textbf{K})^T}{\sqrt{d}} ) (\mathbf{x}^{(i)} \mathbf{V})
  \right\}^N_{i=1},
\end{split}
\end{equation}
where  $\textbf{x} \in \mathbb{R}^{N \times H \times F}$ and $\textbf{h} \in \mathbb{R}^{N \times H \times d}$ correspond to $N$ sensors. 

As we aim to capture more complex relationships between different timestamps, it is often useful to stack multiple attention layers to increase the model's representation power~\cite{attention_all_you_need}, and each layer's attention uses its own set of projection matrices. The output of an attention layer is fed as the input of the next attention layer. After multiple layers have been staked, we can take the output of the last layer to feed into 
a predictor. The predictor can be any type of neural network which is responsible of forecasting future values $\mathbf{\hat{X}}$. 
%
%
%
Figure~\ref{fig:stagnostic} illustrates the procedure.

\noindent
\textit{Needs of Spatio-Temporal Aware Modeling: }
When modeling multiple time series, related studies~\cite{st-grat,gman,astgcn} use the same set of key, query, and value projection matrices $\{\mathbf{Q}, \mathbf{K}, \mathbf{V}\}$, 
for all $N$ different time series and during different times (cf. the left of Figure~\ref{fig:stawareaaa}). This gives rise to a \emph{spatio-temporal agnostic} model that fails to capture distinct patterns from different time series and dynamic patterns that change across time. A spatio-temporal model is called for, where different projection matrices $\{\mathbf{Q}^{(i)}_t, \mathbf{K}^{(i)}_t, \mathbf{V}^{(i)}_t\}$ are used for different time series $i$ and during different times $t$ (cf. the right of Figure~\ref{fig:stawareaaa}).



\begin{figure}[h]
     \centering
     \includegraphics[width=0.9\linewidth]{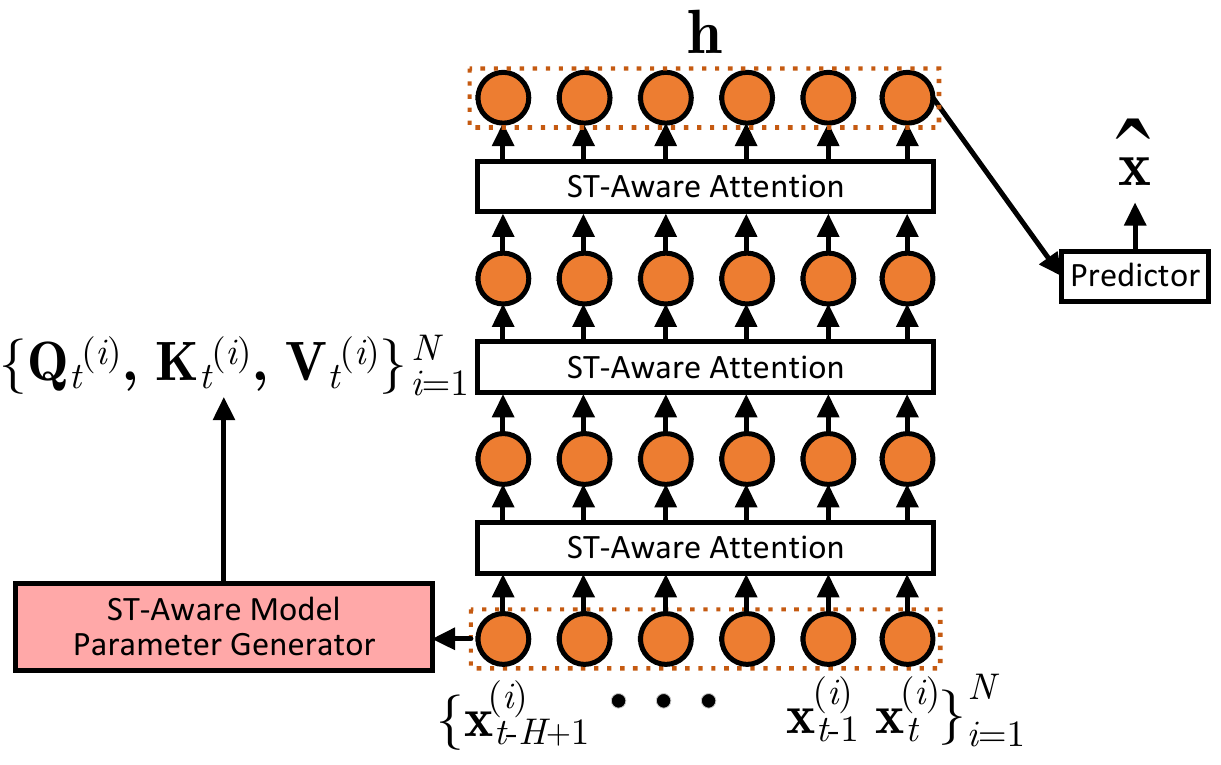}
     \caption{Spatio-temporal Aware Forecasting Model}
     \label{fig:staware}
     \vspace{-10pt}
 \end{figure}

One straightforward solution is to employ a distinct set of projection matrices for each sensor, thus being spatial-aware, and for each time period, thus being temporal-aware. 
%
%
However, this gives rise to a prohibitively large number of parameters to be learned, which leads to high computation time, large memory consumption, slow convergence, and over-fitting issues. Instead, we propose to generate such location-specific and time-varying projection matrices using an encoder-decoder network. The idea is illustrated in Figure~\ref{fig:staware}, where a Spatio-Temporal (ST) Aware Model Parameter Generator generates distinct model parameters $\{\mathbf{Q}^{(i)}_t, \mathbf{K}^{(i)}_t, \mathbf{V}^{(i)}_t\}_{i=1}^{N}$ for $N$ different sensors' time series and during each time window that ends at time $t$. This enables spatio-temporal aware attentions without inuring an explosion of model parameters. 

\section{Methodology}
\label{section:methodology}

\begin{figure*}[ht!]
    \centering
     \centering
     \includegraphics[width=0.85\textwidth]{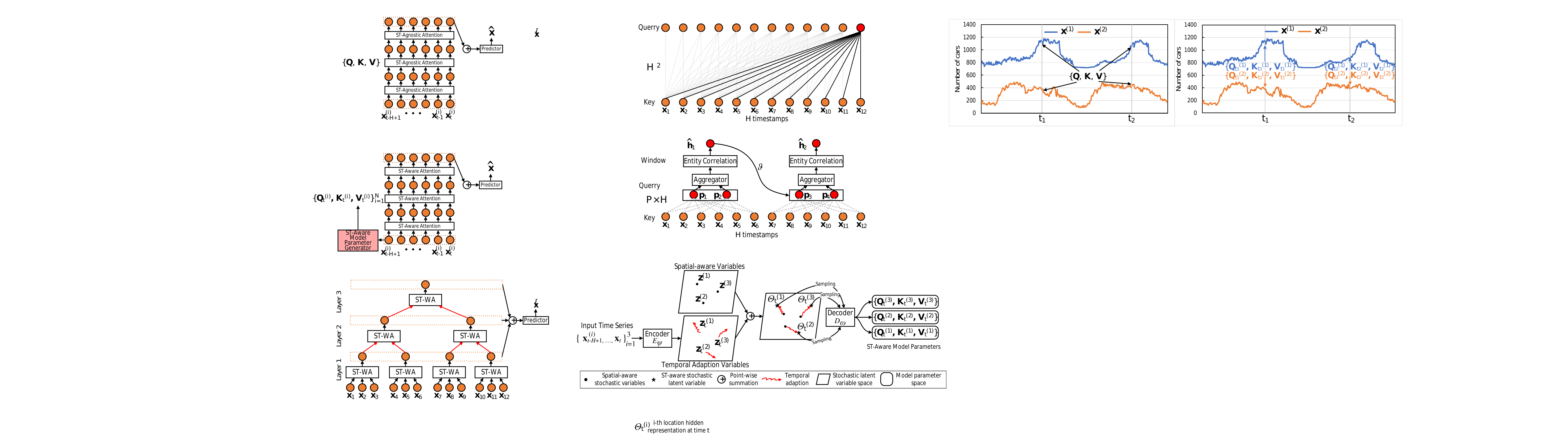}
     \caption{Spatio-temporal Aware Model Parameters Generation}
     \label{fig:weight_generation}
     \vspace{-10pt}
\end{figure*}

\subsection{Spatio-Temporal Aware Modeling}

\subsubsection{Design Considerations}
We propose a \emph{data-driven} and \emph{model-agnostic}  method to generate location-specific and time-varying parameters, thus being able to turn spatio-temporal agnostic models to spatio-temporal aware models. 

First, the method is purely data-driven that only relies on the time series themselves (see the input of the Encoder in Figure~\ref{fig:weight_generation}) and does not rely on any additional information, e.g., points-of-interest around the deployed locations of the sensors, or the knowledge on when morning and afternoon peak hours start at different locations. 

Second, the method is not restricted to a specific type of model 
as the decoder can produce model parameters for different types of models. Figure~\ref{fig:weight_generation} shows an example decoder that outputs location-specific and time-varying model parameters for attentions, i.e., projection matrices. The above two features ensure smooth transformations of existing, different types of spatio-temporal agnostic models to spatio-temporal aware models without using additional, third party information. 
%

In the following, we use attentions as an example to illustrate how to use the method to turn spatio-temporal agnostic attentions to spatio-temporal aware attentions. In the experiments, in addition to attentions, we also test the method on \texttt{RNNs} to justify the method is model-agnostic.

We proceed to offer an overview of the spatio-temporal aware parameter generation, as shown in Figure~\ref{fig:weight_generation}. 
We aim to learn a unique stochastic latent variable $\mathbf{\Theta}_t^{(i)}$ for each location's time series during each time period. More specifically, $\mathbf{\Theta}_t^{(i)}$ denotes the variable from the $i$-th time series for a period ending at time $t$. We design $\mathbf{\Theta}_t^{(i)}$ as the sum of a spatial-aware variable $\textbf{z}^{(i)}$ and a temporal adaption variable $\textbf{z}^{(i)}_t$. 
Then, we generate location-specific and time-varying model parameters using a decoder from $\mathbf{\Theta}_t^{(i)}$. 
This facilities us to generate spatio-temporal aware parameters for different models, thus being model-agnostic. %
In the following, we first cover the learning of stochastic latent variable $\mathbf{\Theta}_t^{(i)}$ and then model parameter decoding from $\mathbf{\Theta}_t^{(i)}$. 

\begin{algorithm}
\caption{Efficient ST-Aware Modeling Algorithm}\label{alg:spatial_temporal_aware}
\SetKwInOut{input}{Input}
\input{\textbf{X}}
\small{
$\small{\left\{\boldsymbol{\mu}^{(i)}\right\}_{i=1}^N}, \small{\left\{\boldsymbol{\Sigma}^{(i)}\right\}_{i=1}^N} \leftarrow $ randomly initialize\;
$\small{E_\psi, D_\omega, \left\{\textbf{P}^{(i)}\right\}_{i=1}^N}  \leftarrow $ randomly initialize\;
\While{did not converge}{
Sample a batch of training data and labels from \textbf{X}\;
Sample $\small{\left\{\textbf{z}^{(i)}\right\}_{i=1}^N}$ using Eq. \ref{eq:ed2}\;
Compute $\small{\left\{\boldsymbol{\mu}^{(i)}_t\right\}_{i=1}^N}, \small{\left\{\boldsymbol{\Sigma}^{(i)}_t\right\}_{i=1}^N}$ using Eq. \ref{eq:ed3}\;
Sample $\small{\left\{\textbf{z}^{(i)}_t\right\}_{i=1}^N}$ using Eq. \ref{eq:ed4}\;
$\small{\left\{\mathbf{\Theta}^{(i)}_t\right\}_{i=1}^N} = \small{\left\{\textbf{z}^{(i)}\right\}_{i=1}^N} +  \small{\left\{\textbf{z}^{(i)}_t\right\}_{i=1}^N}$\;
Compute ST-Aware parameters $\small{\left\{\mathbf{\phi}^{(i)}_t\right\}_{i=1}^N}$ using Eq. \ref{eq:de}\;
\ForEach{$w$ in W}{
Update the proxy tensor with information from the previous window using Eq. \ref{eq:recurrent}\;
Compute proxy values $\small{\left\{\textbf{h}^{(i)}_w\right\}_{i=1}^N}$ using Eq. \ref{eq:attention_window_node}\;
Compute proxy weights $\small{\left\{\textbf{A}^{(i)}_w\right\}_{i=1}^N}$ using Eq. \ref{eq:proxy_weight}\;
Aggregate all proxies into $\small{\left\{\hat{\textbf{h}}^{(i)}_w\right\}_{i=1}^N}$  using Eq. \ref{eq:weight_network}\;
}
Optimize parameters using Eq. \ref{eq:loss}\;}
}
\end{algorithm}

\subsubsection{Learning Stochastic Latent Variables $\mathbf{\Theta}_t^{(i)}$}
We design the unique stochastic latent variable $\mathbf{\Theta}_t^{(i)}$
for the $i$-th time series during a time period ending at $t$ as the sum of two stochastic latent variables---spatial-aware variable $\textbf{z}^{(i)}$ and temporal adaption variable $\textbf{z}_t^{(i)}$ (cf. Equation~\ref{eq:ed1}). 
\begin{equation}
\mathbf{\Theta}^{(i)}_t = \textbf{z}^{(i)} +  \textbf{z}^{(i)}_t\label{eq:ed1}
\end{equation}
Here, we expect the spatial-aware variable $\textbf{z}^{(i)}$ to capture the most general and prominent patterns of the $i$-th time series and the temporal adaption variable $\textbf{z}_t^{(i)}$ to accommodate specific variations w.r.t. the most general patterns at different times. 
We use stochastic variables because they generalize better and have stronger representational power~\cite{kingma2013auto} when compared with deterministic variables, which are a special case of stochastic variables with covariance matrix being all 0.  


The spatial-aware stochastic latent variable $\textbf{z}^{(i)}$ is expected to represent the most general and prominent patterns of the $i$-th time series.  It is spatial-aware, as time series from different locations have different variables $\textbf{z}^{(i)}$, and thus are expected to capture different patterns from different locations. 
 
 
\revision{Since we do not rely on any prior knowledge of the distributions of the latent variables, and many real world processes follow Gaussian distributions~\cite{anderson1962introduction}, we assume they  follows a multivariate Gaussian distribution in a $k$-dimensional space (cf. Equation~\ref{eq:ed2}). Moreover, this assumption is well used and accepted in stochastic models such as \cite{kingma2013auto,gaussian2}. } 
In addition, multivariate Gaussian distributions offer nice properties such as analytical evaluation of the KL divergence in the loss function (cf.
Equation~\ref{eq:loss}) and the reparameterization 
trick for efficient gradient computation. \revision{Furthermore, we provide empirical evidence in Table \ref{table:results2} which shows that our proposal outperforms its deterministic variant}.

\begin{equation}
\mathbf{z}^{(i)} \sim \mathcal{N}(\boldsymbol{\mu}^{(i)},\small{\boldsymbol{\Sigma}}^{(i)})\label{eq:ed2}
\end{equation}


%
We let $\textbf{z}^{(i)}$ be directly learnable in a pure data-driven manner, without using an encoder. This means that the mean $\boldsymbol{\mu}^{(i)}$ and covariance matrix ${\boldsymbol{\Sigma}}^{(i)}$ are learnable parameters. 
Alternatively, we may use an encoder to generate $\textbf{z}^{(i)}$ using location related features, such as POIs around the censor locations~\cite{KDD_urban_traffic}. However, such information is often unavailable from time series data itself, but from third parties, which limits the applicability of the method. We thus choose a pure data-drive design. 



Next, the temporal adaption variable  $\textbf{z}_t^{(i)}$ represents changes w.r.t. 
the most prominent patterns captured by $\textbf{z}^{(i)}$ at a particular time $t$, which accommodates the pattern changes during at $t$. Thus, $\textbf{z}_t^{(i)}$ should be temporal aware as it changes over time. To do so, we produce the temporal adaption variable $\textbf{z}^{(i)}_t$ conditioned on the most recent $H$ timestamps from the $i$-th time series.  
\revision{To this end, given $\{\textbf{x}^{(i)}_{t-H+1}, \ldots, \textbf{x}^{(i)}_t\}$ consisting of the most recent $H$ timestamps from the $i$-th time series, inspired by \cite{kingma2013auto} we opt for a variational encoder $E_\psi$, parameterized by $\psi$, e.g., a neural network, to generate the stochastic variable $\textbf{z}^{(i)}_t$,  which we assume also follows a multivariate Gaussian distribution (cf. Equations~\ref{eq:ed3} and~\ref{eq:ed4}). We opted for a variational encoder to generate model parameters due to its ability to capture the distributions of the input data, which better generalize to cases that were not seen during training. } 

%

\begin{align}
\boldsymbol{\mu}^{(i)}_t, \boldsymbol{\Sigma}^{(i)}_t  &= E_\psi (\textbf{x}^{(i)}_{t-H+1}, \ldots, \textbf{x}^{(i)}_t)\label{eq:ed3} \\
\textbf{z}_t^{(i)} &\sim \mathcal{N}(\boldsymbol{\mu}^{(i)}_t, \small{\boldsymbol{\Sigma}}^{(i)}_t)\label{eq:ed4}
\end{align}
Note that $\boldsymbol{\mu}^{(i)}_t$ and $\small{\boldsymbol{\Sigma}}^{(i)}_t$ are the output of the encoder $E_\psi$, where the encoder parameters in $\psi$ are learnable. This is different from the learning of $\textbf{z}^{(i)}$ where  $\boldsymbol{\mu}^{(i)}$ and $\small{\boldsymbol{\Sigma}}^{(i)}$ are directly learnable. 

The two stochastic variables $\textbf{z}^{(i)}$ and $\textbf{z}^{(i)}_t$ need to be in the same $k$-dimensional space, 
%
%
as illustrated by the diamond-shape space in Figure~\ref{fig:weight_generation}, such that the sum of them is meaningful, i.e., still in the same space. 

\subsubsection{Decoding to Spatio-Temporal Aware Model Parameters}
We use a decoder $D_\omega$ parameterized by $\omega$, e.g., a neural network, to decode the learned stochastic latent variable $\mathbf{\Theta}^{(i)}_t$ to model parameters $\phi^{(i)}_t$ for a specific type of forecasting model, e.g., attentions, or \texttt{RNNs}, as shown in Equation~\ref{eq:de}. 
Since we can not directly backpropagate the gradients through a distribution, which is required to achieve an end to end training, we use the reparameterization trick \cite{kingma2013auto} to circumvent this issue.
%
%
%
More specifically, we first obtain samples according to the distribution specified by $\mathbf{\Theta}^{(i)}_t$, which are then used as the input to the decoder to generate model parameters (cf. Equation~\ref{eq:de}). 
%

\begin{equation} 
\begin{split}
\mathbf{\phi}_t^{(i)} &=
D_\omega(\mathbf{\Theta}_t^{(i)}) \; \; \; 
\label{eq:de}
\end{split}
\end{equation}

Figure~\ref{fig:weight_generation} shows an example where the decoder maps the stochastic latent variables $\mathbf{\Theta}_t^{(i)}$ into model parameters for attention, i.e., projection matrices, where we have  $\phi^{(i)}_t=\{\mathbf{Q}_t^{(i)}, \mathbf{K}_t^{(i)}, \mathbf{V}_t^{(i)}\}$. 
When using the obtained location-specific, time-varying projection matrices in attentions, it turns spatio-temporal-agnostic attention into spatio-temporal-aware attention, as shown in Equation~\ref{eq:attention_st}.

\vspace{-5pt}
\begin{equation} \label{eq:attention_st}
 \begin{split}
  \textbf{h} &= \mathcal{A}tt (\textbf{x} | 
  \phi^{(1)}_t, \ldots, \phi^{(N)}_t) 
  \\ &=  \left\{ 
 \sigma  (\frac{ ( \mathbf{x}^{(i)}  \mathbf{Q}^{(i)}_{t})   ( \mathbf{x}^{(i)} \textbf{K}^{(i)}_{t})^T}{\sqrt{d}} ) (\mathbf{x}^{(i)} \mathbf{V}^{(i)}_{t})
  \right\}^N_{i=1}
\end{split}
\end{equation}
Figure~\ref{fig:staware} shows an overview of the forecasting model based on spatio-temporal aware attentions. 
\begin{figure*}[ht!]
\hfill
     \centering
     \begin{subfigure}[b]{0.495\textwidth}
         \centering
         \includegraphics[width=1.0\textwidth]{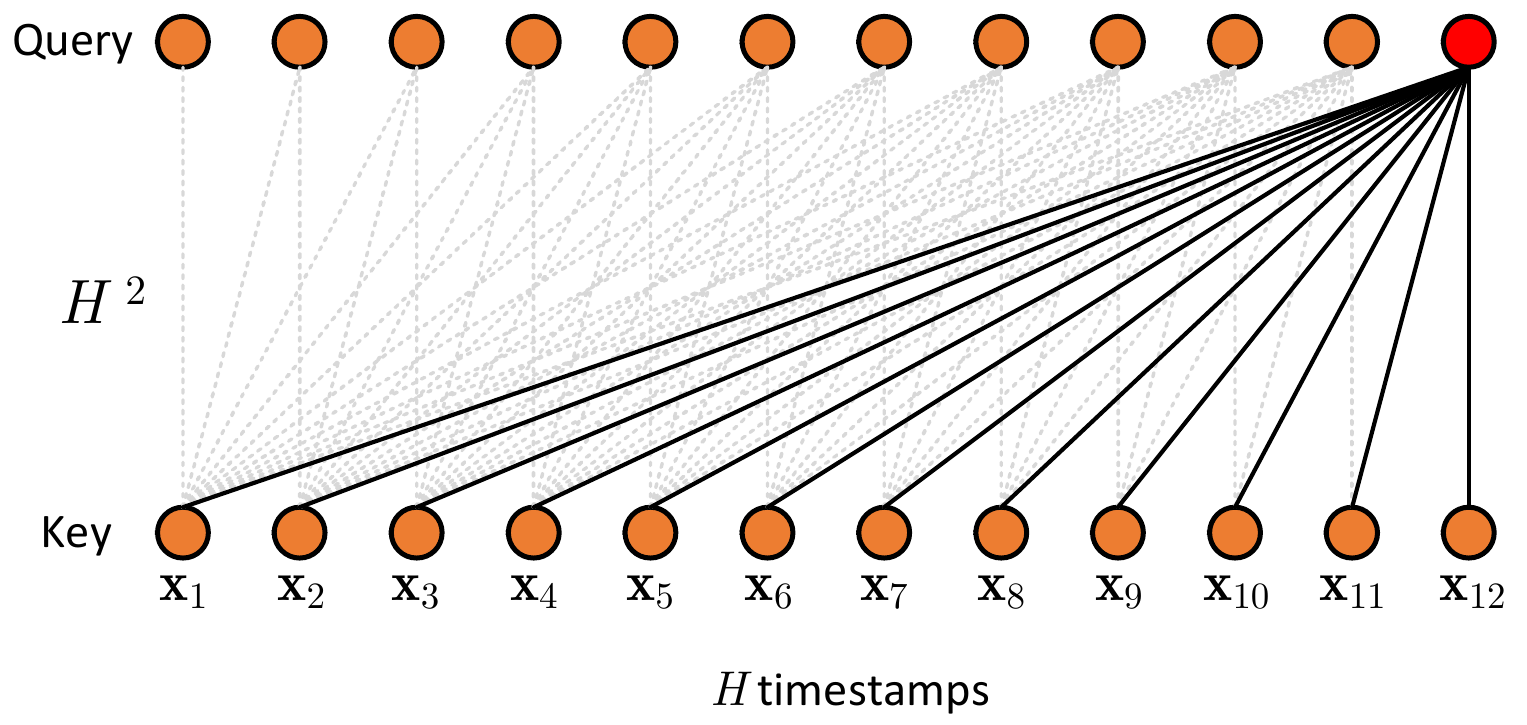}
         \caption{Canonical Attention, $\Theta(H^2)$}
         \label{fig:CA}
     \end{subfigure}
     \begin{subfigure}[b]{0.495\textwidth}
         \centering
         \includegraphics[width=1.0\textwidth]{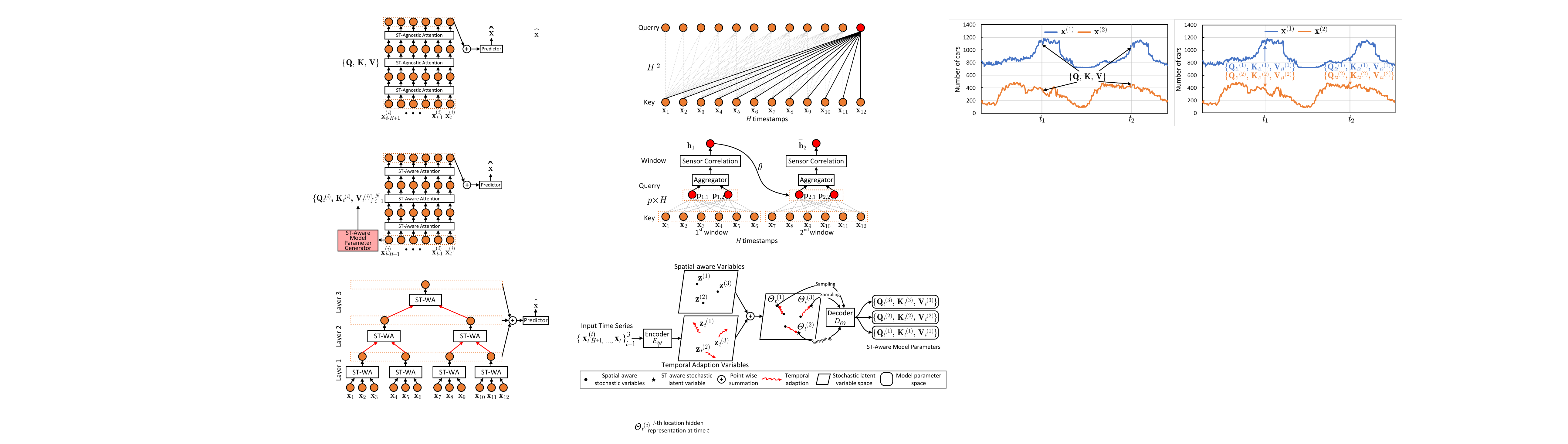}
         \caption{Window Attention, $\Theta(H)$}
         \label{fig:PA}
     \end{subfigure}
     \hfill
     \captionsetup{justification=justified}
     \vspace{-12pt}
     \caption{Complexity Comparison between Canonical Attention (\texttt{CA}) vs. Window Attention (\texttt{WA}). In \texttt{CA}, both Key and Query include all $H$ timestamps as the input time series, and each timestamp in Key attends to all timestamps in Query. In \texttt{WA}, we split the input time series into multiple windows (2 windows in this example) and we perform attentions per window. For each window, Query only includes a limited number of $p$ proxies ($p=2$ in this example) and each timestamp in Key attends to each proxy in Query.}
      \vspace{-12pt}
     \label{fig:PAlinear}
\end{figure*}

Comparing to the naive approach  that directly learn projection matrices for each location,  the proposed spatio-temporal aware model parameter generation reduces significantly the amount of parameters to be learned. In the naive way where each of the $N$ sensors has its own set of projection matrices. This gives rise to $\mathcal{O}(N \times d^2)$ parameters. Such a high number of parameters require high memory consumption and also lead to over-fitting and poor scalability w.r.t. the number of sensors. In contrast, our proposal requires $\mathcal{O}(N \times k)$ parameters for the spatial stochastic latent variable $\boldsymbol{\mu}$ and $\boldsymbol{\Sigma}$. In addition when using a 2 layer MLP network as the decoder $D_\omega$ with $m_1$ and $m_2$ neurons then we only need $\mathcal{O}(k \times m_1 + m_1 \times m_2 + m_2 \times d^2)$ as the decoder is shared across all sensors. 
By doing so we decouple the number of sensors $N$ from $d^2$ which is the dominant term. 
Furthermore, now $m_1$ and $m_2$ are hyper-parameters which can be controlled such that it offers a trade-off between number of parameters, i.e. memory usage, and efficiency.

\subsection{Efficient Spatio-Temporal Aware Attention} 

Although attention based models 
show promising accuracy, a limitation is its quadratic complexity $\mathcal{O}(H^2)$, where $H$ is the size of the input time series. This is because each timestamp in the Key attends to all timestamps in the Query (cf. Figure~\ref{fig:CA}). When the input time series is long, e.g., in settings of long term forecasting, the quadratic complexity becomes a serious barrier. 

The spatio-temporal aware parameter generation incurs further computation costs, making its efficiency even worse. To contend with the efficiency challenges, we propose an efficient \emph{Window Attention} (\texttt{WA}) with linear complexity, to ensure competitive overall efficiency, even when using the spatio-temporal aware parameter generation. 

We propose to break down the input time series of size $H$ into $W = H/S$ smaller windows along the temporal dimension where $S$ is the window size. We compute attention scores per window; and 
for each window, we introduce multiple (a small constant, e.g., 2 or 3) learnable proxies which act as the ``Query'' in the canonical attention. Within a window, each timestamp in the Key only computes a single attention score w.r.t. each proxy in the Query, thus achieving linear complexity. 

More specifically, in canonical attention, each of the $S$ timestamps in a window should compute an attention score w.r.t. each timestamp in the windows, thus having $O(S^2)$. Instead, assuming that we have $p$ proxies in window attentions, each of the $S$ timestamps in the window computes an attention score w.r.t. each proxy, resulting in $O(p\cdot S)$. Since we use a small constant of proxies, i.e., $p$ being a constant, then we have linear complexity $O(S)$ for window attentions. 

The idea is illustrated in Figure~\ref{fig:PA} with an example input time series of size $H=12$ being split into $W=2$ windows, where the window size is $S=6$ and each window has $p=2$ proxies. 
%
We proceed to describe the details and the design intuition behind proxies. We assume that for each window there exist $p$ representative temporal patterns, which we use $p$ proxies to capture. 
The proxies then replace the Query inside canonical {self-}attention. 
%
To this end, we introduce a distinct learnable proxy tensor $\textbf{P} \in \mathbb{R}^{W \times N \times p \times d}$, such that during each of the $W$ windows, each of the $N$ sensors has a total of $p$ proxies. 
Each proxy uses a $d$-dimensional vector, which 
intends to capture a specific, hidden temporal pattern in the window. 
%

Specifically, when considering the $i$-th time series and the $w$-th window, $\textbf{P}^{(i)}_w \in \mathbb{R}^{p \times d}$ replaces the Query $\textbf{x}^{(i)}\textbf{W}_Q \in \mathbb{R}^{H \times d}$ in Equation~\ref{eq:attention_multiple}. This serves as a set of $p$ proxies. Then, each time timestamp in window $w$ of the $i$-th time series needs to compute a single attention score w.r.t. each proxy of the $p$ proxies in $\textbf{P}^{(i)}_w$. Formally the operation is defined as shown in Equation \ref{eq:attention_window_node}: 


\begin{equation} \label{eq:attention_window_node}
\begin{split}
  \textbf{h}^{(i)}_w &= \mathcal{A}tt (\mathbf{x}_w^{(i)} \;|\; \textbf{P}^{(i)}_w, \textbf{K}_t^{(i)}, \textbf{V}_t^{(i)})\\
  &= 
  \left\{\sigma  (\frac{ \textbf{P}^{(i)}_{w,j}   ( \mathbf{x}_w^{(i)} \textbf{K}_t^{(i)})^T}{\sqrt{d}} ) (\mathbf{x}_w^{(i)} \mathbf{V}_t^{(i)})\right\}_{j=1}^{p},
\end{split}
\end{equation}
where $\textbf{x}_w^{(i)}=(\textbf{x}^{(i)}_{w*S +1},\textbf{x}^{(i)}_{w*S +2},..., \textbf{x}^{(i)}_{(w+1)*S})$ 
and the output of window $w$ for time series $i$ is $\textbf{h}^{(i)}_w \in \mathbb{R}^{p \times d}$. To get the output of the layer we need to 
apply Equation~\ref{eq:attention_window} for each window $w$:

\vspace{-5pt}
\begin{equation} \label{eq:attention_window}
  \textbf{h}^{(i)} = \left\{ \mathcal{A}tt (\textbf{x}^{(i)}_{w} | \textbf{P}^{(i)}_w, \textbf{K}^{(i)}, \textbf{V}^{(i)}) \right\}_{w=0}^{W-1},
\end{equation}
where $\textbf{h}^{(i)} \in \mathbb{R}^{W \times p \times d}$.  
Note that by applying the proxies, we reduce the complexity of canonical self-attention on an input time series of size $H$ from quadratic $\mathcal{O}(H^2)$ to $\mathcal{O}(W \cdot p \cdot S)$. Since the number of proxies $p$ is a constant and since $W\cdot S=H$, 
the asymptotic complexity becomes linear, i.e., $\mathcal{O}(H)$. 

\begin{figure}[H]
     \centering
     \includegraphics[width=0.8\linewidth]{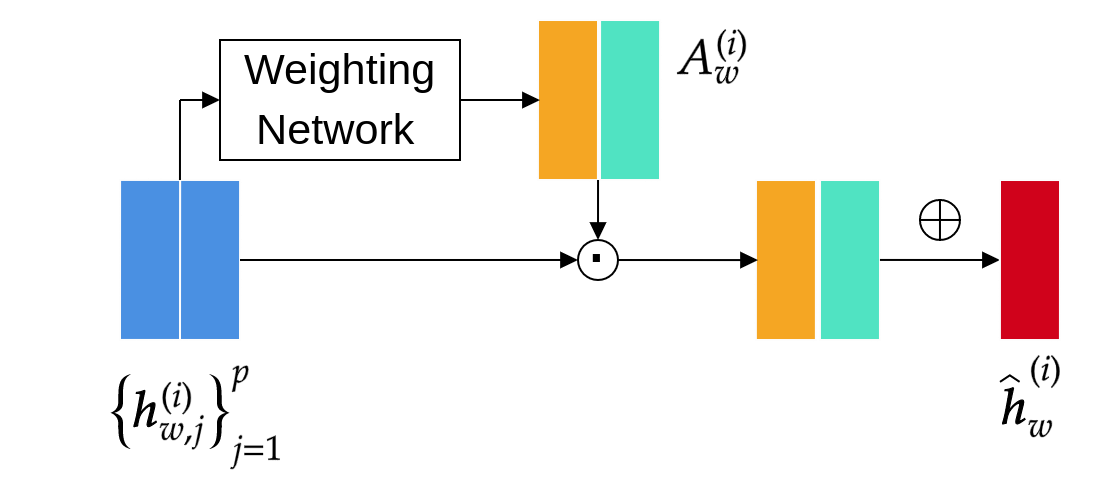}
     \caption{Illustration of Proxy Aggregator}
     \label{fig:aggregator}
     \vspace{-5pt}
\end{figure}

To further reduce the complexity for the upcoming layers, we propose an aggregation function responsible of aggregating \revision{all the $p$ proxies within a window $w$, i.e., $\textbf{h}_w^{(i)}=\{\textbf{h}_{w,j}^{(i)}\}_{j=1}^{p}$, into a single, shared hidden representation $\hat{\textbf{h}}_w^{(i)}$. } The resulting $\hat{\textbf{h}}_w^{(i)}$ is shrinked by a factor of $p$, thus giving rise to less attention scores computation at the next layer. The idea is illustrated in Figure \ref{fig:aggregator}. 

First, we use a weighting network, e.g., 2-layer neural network as shown in Equation \ref{eq:proxy_weight}, to weigh each proxy. 

\begin{equation}\label{eq:proxy_weight}
\textbf{A}_w^{(i)} = f_2 ( \textbf{W}_2 ( f_1 (\textbf{W}_1 \textbf{h}_w^{(i)})))
\end{equation}
where $\textbf{W}_1$ and $\textbf{W}_2$ are trainable parameters which are shared among different windows. $f_1$ is a non-linear activation, which is set to $tanh$ in our experiments, and $f_2$ represents the $Sigmoid$ activation function which ensures the weights are bounded in range [0, 1], controlling the information flow. 
Next, after we have computed the proxy weight scores $\textbf{A}_w^{(i)} \in \mathbb{R}^{p \times d}$ we can calculate the final window representation as follows:  

\begin{equation}\label{eq:weight_network}
    \hat{\textbf{h}}_w^{(i)} = \sum^{p}_{j=1} \textbf{A}_{w,j}^{(i)} \odot \textbf{h}_{w,j}^{(i)} 
    \vspace{-5pt}
\end{equation}
where $\odot$ represents the point-wise multiplication. We first scale each proxy $\textbf{h}^{(i)}_p$ with its respective weight coefficient $\textbf{A}^{(i)}_p$ and then sum all the proxies within the same window.

The reduced complexity comes with a price that the temporal receptive filed of each timestamp in window attention is reduced from $H$, as in the canonical attention, to $S$, i.e., the window size $S$. This makes it impossible to capture relationships among different windows and also long-term dependencies, thus adversely affecting accuracy. To compensate for reduced temporal receptive field, we connect the output of a window to the next window proxies such that the temporal information flow is maintained. \revision{More specifically, we utilize a function $\vartheta$, e.g., a neural network, to fuse the output of the previous window $\hat{\textbf{h}}_{w-1}^{(i)}$ with all $p$ learnable proxies $\left\{\textbf{P}_{w,j}^{(i)}\right\}_{j=1}^p$ in the current window before the window attention computation (see $\vartheta$ in Figure~\ref{fig:PA}). 
\revision{
\begin{equation}\label{eq:recurrent}
    \textbf{P}_{w,j}^{(i)} = \vartheta(\hat{\textbf{h}}^{(i)}_{w-1} \; || \; \textbf{P}^{(i)}_{w, j}) \; \; \; \forall \; j \in [1, p]
\end{equation}}
}
\vspace{-1.7em}

\subsection{Sensor Correlation Attention}
Traffic time series from different sensors often show correlations---traffic from one sensor is often influenced by traffic from nearby sensors, thus the traffic information captured by the sensors show correlations. 
%
%
In order to accurately predict the future, we need a way to 
model such correlations, which we refer to as \textit{sensor correlations}.
We proceed to elaborate how we can utilize the same attention mechanism introduced already to capture the interactions between different sensors within a window $w$. After we aggregate the proxies, the output of a window attention is $\hat{\textbf{h}}_w \in \mathbb{R}^{N \times d}$, where the window information is summarized into a fixed representation for the $i$-th sensor, where $1\leq i \leq N$. To capture how one sensor $i$ if affected by another sensor $j$, we use a normalized embedded Gaussian function:

\begin{equation}
    \textbf{B}(\hat{\mathbf{h}}_w^{(i)}, \hat{\mathbf{h}}_w^{(j)}) = \frac{e^{\theta_1 (\hat{\mathbf{h}}_w^{(i)})^\mathsf{T} \theta_2(\hat{\textbf{h}}_w^{(j)})}}{\sum_{j=1}^N{e^{\theta_1 (\hat{\mathbf{h}}_w^{(i)})^\mathsf{T} \theta_2(\hat{\mathbf{h}}_w^{(j)})}}},
    \label{eq:spatial_att}
\end{equation}
where $\theta_{1}$ and $\theta_{2}$ are two embedding functions to perform a linear transformation on the data between sensors $i$ and $j$. The numerator of Equation \ref{eq:spatial_att} calculates the similarity between the source sensor $i$ and the target sensor $j$, while the denominator ensures that the final attention scores for a specific target sensor is normalised. Next, we can get a new updated representation of the $i$-th sensor  $\overline{\textbf{h}}^{(i)} \in \mathbb{R}^{N \times d}$ as shown in Equation \ref{eq:spatial_att2}:   

\begin{equation}\label{eq:spatial_att2}
    \overline{\textbf{h}}^{(i)} = \sum^{N}_{j=1} \textbf{B} (\hat{\textbf{h}}_w^{(i)}, \hat{\textbf{h}}_w^{(j)}) \odot \hat{\textbf{h}}^{(j)},
\end{equation} 
where $\odot$ represents points-wise multiplication. Similar to temporal attention, we assume that a single set of linear transformation $\theta_{1}$ and $\theta_{2}$ might not be sufficient to describe all the interactions between different sensors, and thus we can use the model parameters generation process already introduced in Section \ref{section:methodology} to generate a distinct set of transformation matrices for each sensor. 


\subsection{Full Model}



\begin{figure}[h]
     \centering
     \includegraphics[width=\linewidth]{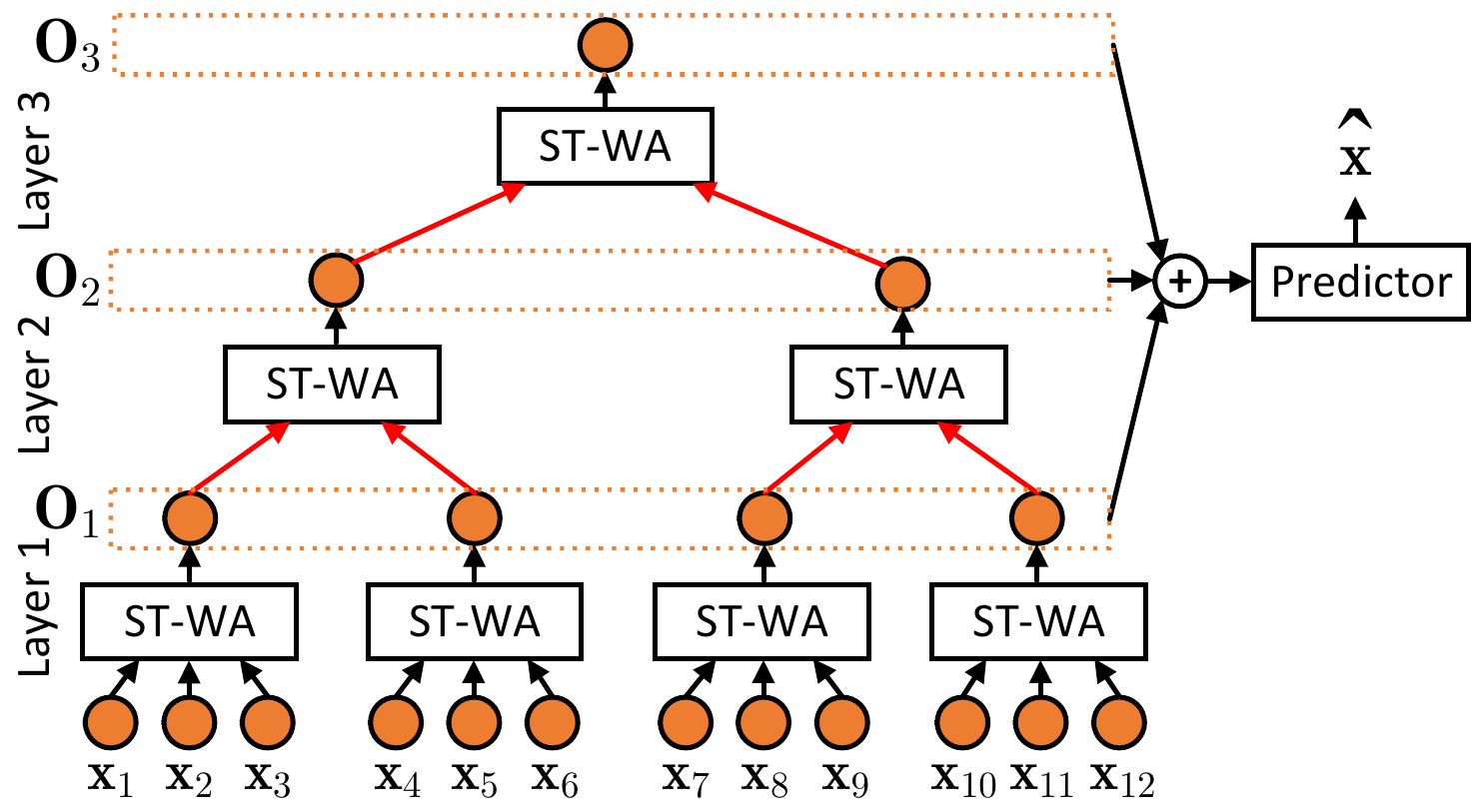}
     \caption{Full Model}
     \label{fig:model} 
     \vspace{-10pt}
\end{figure}

When using attentions, multiple layers of attentions are often stacked together to improve accuracy.
Figure \ref{fig:model} shows the full model, 
%
which consists of multiple layers of 
Spatio-Temporal Aware Window Attention (\texttt{ST-WA}), where location-specific and  time-varying projection matrices are used.
\revision{Each window attention layer has its own set of learnable proxy tensor \textbf{P}}. The output of the $l$-th layer is derived by concatenating the output of each window:

\begin{equation}
    \textbf{O}_l = [\overline{\textbf{h}}_1, \overline{\textbf{h}}_2,...,\overline{\textbf{h}}_W], 
\end{equation}
where  $\textbf{O}_l \in \mathbb{R}^{W \times N \times d}$.  To ease gradient propagation we create gradient shortcuts by connecting each layer output directly to the predictor.  Since the dimension of the input is exponentially decreased at each layer we use a skip-connection in the form of 1 layer neural network to ensure consistent output shape between each layer as shown in Equation \ref{eq:skip_conn}.

\begin{equation}\label{eq:skip_conn}
    \textbf{O} = \sum_{l=1}^{L} \textbf{W}_l (\textbf{O}_l),
\vspace{-3pt}
\end{equation}
where $L$ represents the total number of layers, and $\textbf{W}_{l}$ represents the learnable parameters of the skip connection at layer $l$. Finally we employ a predictor with a 2 layer neural network which is responsible of mapping the hidden representation \textbf{O} to future predictions $\hat{\textbf{X}}$ as shown in Equation \ref{eq:predictor}.

\begin{equation}\label{eq:predictor}
    \hat{\textbf{X}} = \textbf{W}_4 (ReLU (\textbf{W}_3 (\textbf{O}))),
\end{equation}
where $ReLU$ represents the rectified linear unit activation function, $\textbf{W}_3$ and $\textbf{W}_4$ are learnable weight matrices. 

\noindent
\textbf{Complexity Analysis}
When using attention, multiple layers of attention are often stacked together to improve accuracy. 
In canonical self-attentions, each attention layer has the same input size, e.g., the input time series size $H$. Thus, when having $L$ layers, the complexity is $\Theta(LH^2)$. 

When using window attention, the size of $(l+1)$-th layer is only $\displaystyle\frac{1}{S_l}$ of the size of the $l$-th layer, where $S_l$ is the window size of the $l$-th layer. 
Consider the example shown in Figure~\ref{fig:model} with 3 layers of window attention (\texttt{WA}). 
The size of the input time series is $H=12$. The window size in the first layer $S_1$ is 3, making the input of the 2-nd layer \texttt{WA} to be $\displaystyle\frac{1}{3}\cdot 12=4$. As the 2-nd layer's window size $S_2$ is 2, the input of the 3-rd layer \texttt{WA} has an input size of  $\displaystyle\frac{1}{2}\cdot 4=2$. Thus, the input size of the $L$-th layer is at most $\displaystyle\frac{H}{\hat{S}^{L-1}}$, where $\hat{S}=\min_{1\leq i \leq L} S_i$ is the minimum window size of all layers. Since $\displaystyle\sum_{i=1}^{L} \frac{H}{\hat{S}^{i-1}} < \frac{\hat{S}}{\hat{S}-1}\cdot H$,  
the complexity of $L$ stacks of \texttt{WA} is still linear $\mathcal{O}(H)$.

\subsection{Loss Function and Optimization}
We train the proposed spatio-temporal aware parameters generation by optimizing the loss function shown in Equation~\ref{eq:loss} in an end-to-end fashion. 

\begin{equation}\label{eq:loss}
\begin{split}
    \operatorname*{argmin}_{\psi,\omega, \{\boldsymbol{\mu}_i\}, \{\boldsymbol{\Sigma}_i\}} \sum_{t}\mathcal{H}(&\mathbf{X}_{t+1,\ldots, t+U}, \hat{\mathbf{X}}_{t+1,\ldots, t+U}) \\ + &\alpha D_{KL}[ \Theta_t \: || \: \hat{p}]
\end{split}
\end{equation}
The first term measures the discrepancies between the predictions $\hat{\textbf{X}}_{t+1,\ldots, t+U}$ and the ground truth values $\textbf{X}_{t+1,\ldots, t+U}$. We use Huber loss to measure the discrepancies in our experiments, since it is less sensitive to outliers in the training data~\cite{huber1992robust}.

\begin{equation}
\small
     \mathcal{H}(\mathbf{X}_U, \hat{\mathbf{X}}_U) = 
 \begin{cases} 
  \displaystyle\frac{1}{2}(\mathbf{X}_U - \mathbf{\hat{X}})_U^2 \quad \quad \quad  \;|\mathbf{X}_U - \mathbf{\hat{X}}_U| \leq \delta \\ 
 \delta (|\mathbf{X}_U - \mathbf{\hat{X}}_U| - \displaystyle\frac{1}{2} \delta)   \;\;\; \text{otherwise} 
 \end{cases} 
\end{equation}

Here, $\delta$ is a threshold parameter that controls the range of squared error loss. The second term  is a regularizer w.r.t. a prior distribution $\hat{p} = \mathcal{N}(0, \textbf{I})$.

Different from traditional learning settings, where we directly optimize the model parameters, e.g., projection matrices in attention. In our setting, the model parameters are generated from a stochastic latent variable by a decoder. Thus, the learning goal is to optimize the parameters that generate the model parameters. More specifically, the decoder $D_\omega$ is parameterized by $\omega$; the stochastic hidden variable $\mathbf{\Theta}_t$ is the sum of the location specific stochastic variable decided by the learnable parameters $\{\boldsymbol\mu_i\}_{i=1}^N$ and $\{\boldsymbol\Sigma_i\}_{i=1}^N$; and the stochastic temporal adaption that is returned by the encoder $E_\psi$ parameterized by $\psi$. Thus, minimizing the loss function enables the identification of the optimal parameters $\omega$, $\psi$, $\{\boldsymbol\mu_i\}_{i=1}^N$ and $\{\boldsymbol\Sigma_i\}_{i=1}^N$. 
When learning the stochastic latent variable $\boldsymbol{\Theta}_t$, in order to enforce our assumption that it follows a multivariate Gaussian distribution, we 
 make the learned posterior distribution close to a prior distribution
$\hat{p} = \mathcal{N}(0, \textbf{I})$. 
Here, $D_{KL}$ represents the Kullback–Leibler divergence and $\alpha$ is a penalty term to control the contribution of the regularization term. We empirically study the impact of the regularizer and reported the results in the Section \ref{section:results}.

\section{Experimental Studies}\label{section:experiments}

\subsection{Experimental Setup} 
\label{ssec:expsetup}
\textbf{Data Sets:} We use four commonly used, public traffic time series data sets~\cite{stsgcn,agcrn,stfgnn}. The data has been collected by California Transportation Agencies Performance Measurement System (PEMS, \url{http://pems.dot.ca.gov/}) and is released by~\cite{stsgcn}.  Every 5 minutes, a time series has $F=1$ attribute, representing the average traffic flow. Thus, no personally identifiable information, e.g., information on individual vehicles, is revealed. To facilitate fair comparisons, we follow existing studies~\cite{astgcn,agcrn} by splitting the data sets in chronological order with 60\% for training, 20\% for validation, and 20\% for testing.  The number of time series $N$ and the duration of each dataset is shown in Table~ \ref{table:results}.

\begin{table*}[h!]
\renewcommand{\tabcolsep}{2.2pt}

\centering

\begin{tabular}{|c|c|cccccccc|cc|c|c|} 
\hline
\multicolumn{2}{|c|}{\begin{tabular}[c]{@{}c@{}}\\Baseline methods\end{tabular}}                             & \multicolumn{8}{c|}{ST-agnostic}                                                                                                                                    & \multicolumn{2}{c|}{S-aware}                                         & \begin{tabular}[c]{@{}c@{}}T-aware\\\end{tabular} & ST-aware          \\ 
\hline
Dataset                                                                                             & Metric & $\texttt{LongFormer}$ & $\texttt{DCRNN}$ & $\texttt{STGCN}$ & $\texttt{STG2Seq}$ & $\texttt{GWN}$ & $\texttt{STSGCN}$ & $\texttt{ASTGNN}$       & $\texttt{STFGNN}$ & $\texttt{EnhanceNet}$                             & $\texttt{AGCRN}$ & $\texttt{meta-LSTM}$                              & $\texttt{ST-WA}$  \\ 
\hline
\multirow{3}{*}{\begin{tabular}[c]{@{}c@{}}\textbf{~PEMS03~}\\\textbf{~}N=358\\~T=3mo\end{tabular}} & MAE    & 17.50                 & 18.18            & 17.49            & 19.03              & 19.85          & 17.48             & \textbf{\textbf{15.07}} & 16.77             & $16.05$                                           & $16.06$          & 19.89                                             & 15.17             \\
                                                                                                    & MAPE   & 16.80                 & 18.91            & 17.15            & 21.55              & 19.31          & 16.78             & \textbf{\textbf{15.80}} & 16.30             & $15.83$                                           & $15.85$          & 20.38                                             & 15.83             \\
                                                                                                    & RMSE   & 30.24                 & 30.31            & 30.12            & 29.73              & 32.94          & 29.21             & 26.88                   & 28.34             & $28.33$                                           & $28.49$          & 33.71                                             & \textbf{26.63 }   \\ 
\hline
\multirow{3}{*}{\begin{tabular}[c]{@{}c@{}}\textbf{PEMS04}\\N=308\\~T=2mo\textbf{}\end{tabular}}    & MAE    & 23.83                 & 24.70            & 22.70            & 25.20              & 25.45          & 21.19             & 19.26                   & 19.83             & \begin{tabular}[c]{@{}c@{}}$20.44$\\\end{tabular} & 19.83            & 25.37                                             & \textbf{19.06 }   \\
                                                                                                    & MAPE   & 15.57                 & 17.12            & 14.59            & 18.77              & 17.29          & 13.90             & 12.65                   & 13.02             & $13.58$                                           & 12.97            & 17.09                                             & \textbf{12.52 }   \\
                                                                                                    & RMSE   & 37.19                 & 38.12            & 35.55            & 38.48              & 39.70          & 33.65             & 31.16                   & 31.88             & $32.37$                                           & 32.26            & 39.90                                             & \textbf{31.02 }   \\ 
\hline
\multirow{3}{*}{\begin{tabular}[c]{@{}c@{}}\textbf{PEMS07}\\N=883\\~T=4mo\textbf{}\end{tabular}}    & MAE    & 26.80                 & 25.30            & 25.38            & 32.77              & 26.85          & 24.26             & 22.23                   & 22.07             & $21.87$                                           & $21.29$          & 27.02                                             & \textbf{20.74 }   \\
                                                                                                    & MAPE   & 12.11                 & 11.66            & 11.08            & 20.16              & 12.12          & 10.21             & 9.25                    & 9.21              & $9.13$                                            & $8.97$           & 12.32                                             & \textbf{8.77 }    \\
                                                                                                    & RMSE   & 42.95                 & 38.58            & 38.78            & 47.16              & 42.78          & 39.03             & 35.95                   & 35.80             & $35.57$                                           & $35.12$          & 43.00                                             & \textbf{34.05 }   \\ 
\hline
\multirow{3}{*}{\begin{tabular}[c]{@{}c@{}}\textbf{PEMS08}\\N=170\\~T=2mo\textbf{}\end{tabular}}    & MAE    & 18.52                 & 17.86            & 18.02            & 20.17              & 19.13          & 17.13             & 15.98                   & 16.64             & $16.33$                                           & 15.95            & 19.99                                             & \textbf{15.41}    \\
                                                                                                    & MAPE   & 13.66                 & 11.45            & 11.40            & 17.32              & 12.68          & 10.96             & 9.97                    & 10.60             & $10.39$                                           & 10.09            & 12.51                                             & \textbf{9.94 }    \\
                                                                                                    & RMSE   & 28.68                 & 27.83            & 27.83            & 30.71              & 31.05          & 26.80             & 25.67                   & 26.22             & $25.46$                                           & 25.22            & 31.65                                             & \textbf{24.62}    \\
\hline
\end{tabular}
\caption{Overall Accuracy, $H=12$, $U=12$}
\label{table:results}
\vspace{-15pt}
\end{table*}

\noindent
\textbf{Forecasting setting:} We consider a commonly used forecasting setup as the default setup~\cite{stsgcn,dcrnn,graphwavenet}, where given the previous $H=12$ timestamps (1 hour), we predict the following $U=12$ timestamps (1 hour). We consider two additional setups. First, we increase the historical timestamps to $H=36$ (3 hours) and $H=120$ (10 hours) while keeping $U=12$ to study the impact on long term dependency. Second, we increase the  historical timestamps to $H=72$ (6 hours) and the forecasting horizon to $U=72$ (6 hours) to quantify the scalability of the methods. 

\noindent
\textbf{Baselines:} To evaluate the performance of our proposal, we compare with widely used baselines and state-of-the-art models for traffic time series forecasting. In particular, this includes ST-agnostic models (\revision{\texttt{LongFormer}}, \texttt{DCRNN}, \texttt{STGCN}, \texttt{STG2Seq}, \texttt{GWN}, \texttt{STSGCN}, \revision{\texttt{ASTGNN}}, \texttt{STFGNN}), S-aware models (\texttt{EnhanceNet}, \texttt{AGCRN}), and T-awre models (\texttt{meta-LSTM}).
\begin{itemize}
    \item \revision{\texttt{LongFormer}~\cite{beltagy2020longformer} a Transformer like architecture which employs sliding window attention.}
    \item \texttt{DCRNN}~\cite{dcrnn} employs \texttt{GRU} to model temporal dependencies (\emph{TD}) and diffusion graph convolution to model sensor correlations (\emph{SC}). 
    \item  \texttt{STGCN}~\cite{astgcn} uses 2D convolution to model \emph{TD} and graph convolution to model \emph{SC}. 
    \item \texttt{STG2Seq} \cite{bai2019stg2seq} uses gated residual with attentions to model \emph{TD} and  graph convolution to model \emph{SC}. 
    \item \texttt{GWN}~\cite{graphwavenet} Graph WaveNet uses  dilated causal convolutions to model \emph{TD} 
    and graph convolution for \emph{SC}. 
    \item \texttt{STSGCN}~\cite{stsgcn}  a state-of-the-art method which jointly captures \emph{TD} and \emph{SC} using graph convolution.
    \item \revision{\texttt{ASTGNN}~\cite{astgnn} an  encoder-decoder network which incorporates local context into self-attention.}
    \item \texttt{STFGNN} \cite{stfgnn} relies on gated convolutions to jointly capture \emph{TD} and \emph{SC}. 
    \item \texttt{EnhanceNet} \cite{enhancenet} A state-of-the-art spatial-aware method that enhances existing \texttt{RNN} and \texttt{TCN} based models for \emph{TD} and graph convolution for \emph{SC}. 
    \item \texttt{AGCRN} \cite{agcrn} A state-of-the-art spatial-aware method that uses \texttt{RNN} for modeling \emph{TD} and graph convolution for \emph{SC}, 
    achieving the best accuracy so far.
    \item \texttt{meta-LSTM}~\cite{chen2018meta} a temporal-aware  method that uses 
    $\texttt{LSTM}$ to capture \emph{TD}, while \emph{SC} is not captured. 
\end{itemize}
We do not include another spatial-aware model $\texttt{ST-MetaNet}$~\cite{KDD_urban_traffic} as it requires POI information. However, no POI information is available along with the data sets and all other baselines do not rely on POI information. 

\noindent
\textbf{Evaluation Metrics:} For accuracy, we follow existing literature~\cite{dcrnn,graphwavenet} to report mean absolute error (\textit{MAE}), root mean square error (\textit{RMSE}), and mean absolute percentage error (\textit{MAPE}). We also report runtime and memory usage. 

\noindent
\textbf{Implementation Details and Hyperparameter Settings:} We train our model using Adam optimizer with a fixed learning rate of 0.001 and with a batch size of 64. The total number of epochs is set to 200 and we use early stopping with a patience of 15. We tune the hyper-parameters on the validation data by grid search. The number of layers is chosen from $L \in \{1,2,3\}$, window size $w \in \{2,3,6\}$, number of proxies $p \in \{1,2,3\}$, hidden representation $d \in  \{16,32\}$, stochastic latent variable size $k \in \{4,8,16,32\}$.

For $H=12$  we have a default setting in which we stack 3 layers, with $p=1$ and we set the window size $S$ in the 1st, 2nd, and 3rd layers to be $3$, $2$, and $2$, respectively. For $H=72$ we use 3 layers with $p=2$ and $S=6$ across all layers. We set the hidden representation $d=32$. We initialize each sensor's stochastic latent variable $\textbf{z}^{(i)}$ as a $k=16$ dimensional Gaussian distribution using randomly initialized parameters $\mu^{(i)}$ and $\Sigma^{(i)}$. 
We use a 3 layer fully connected network with 32 neurons and $ReLU$ as the activation for the encoder $E_\psi$ to generate a 16-dimensional Gaussian distribution $\textbf{z}^{(i)}_t$ parameterized by $\mu^{(i)}_t$ and $\Sigma^{(i)}_t$. We enforce $\Sigma^{(i)}$ and $\Sigma^{(i)}_t$ to be diagonal matrices. We use a 3 layer fully-connected network with 16, 32, and 5 neurons and $ReLU$ as activation functions for the first 2 layers as the decoder $D_\omega$. For the predictor we use 2 fully connected layers, each with 512 neurons, and $ReLU$ as activation function.  When computing the attentions, to learn different types of relationships, we utilize multi-head attention with a total of 8 heads.

We implement $\texttt{ST-WA}$  and other baselines on Python 3.7.3 using PyTorch 1.7.0. The experiments are conducted on a computer node on an internal cloud, running Ubuntu 16.04.6 LTS, with one Intel(R) Xeon(R) CPU @ 2.50GHz with one Tesla V100 GPU card.  \revision{The code is publicly available on GitHub at
https://github.com/razvanc92/ST-WA.}


\begin{table*}[ht!]
\setlength\tabcolsep{2pt}
\centering

\begin{tabular}{|c|cccc|cccc|cccc|} 
\toprule
\multirow{2}{*}{} & \multicolumn{4}{c|}{$H=12$}                                                     & \multicolumn{4}{c|}{$H=36$}                                                            & \multicolumn{4}{c|}{$H=120$}                                                            \\ 
\cline{2-13}
                  & $\texttt{STFGNN}$ & $\texttt{EnhanceNet}$ & $\texttt{AGCRN}$ & $\texttt{ST-WA}$ & $\texttt{STFGNN}$ & $\texttt{EnhanceNet}$ & $\texttt{AGCRN}$ & $\texttt{ST-WA}$        & $\texttt{STFGNN}$ & $\texttt{EnhanceNet}$ & $\texttt{AGCRN}$ & $\texttt{ST-WA}$         \\ 
\cline{1-1}\cmidrule{2-2}\cline{3-4}\cmidrule{5-6}\cline{7-8}\cmidrule{9-10}\cline{11-12}\cmidrule{13-13}
MAE               & 19.83             & 20.44                 & 19.38            & \textbf{19.06}   & 20.93             & 20.26                 & 19.69            & \textbf{\textbf{18.90}} & 19.36             & 19.85                 & 19.87            & \textbf{\textbf{18.90}}  \\
MAPE              & 13.02             & 13.58                 & 12.89            & \textbf{12.52}   & 13.74             & 13.29                 & 12.91            & \textbf{\textbf{12.43}} & 13.20             & 13.53                 & 13.59            & \textbf{12.92}           \\
RMSE              & 31.88             & 32.37                 & 31.29            & \textbf{31.02}   & 32.90             & 32.09                 & 32.27            & \textbf{30.92}          & 30.94             & 31.63                 & 31.65            & \textbf{\textbf{30.69}}  \\
\bottomrule
\end{tabular}

\caption{Impact of $H$, \textbf{PEMS04}}
\label{table:long_range_res}
\end{table*}

\begin{table*}[!htb]
    \begin{minipage}{.5\linewidth}
      \centering
\centering

\begin{tabular}{|c|c|cccc|} 
\toprule
\multicolumn{2}{|c|}{Baseline methods}                                                    & \multirow{2}{*}{$\texttt{STFGNN}$} & \multirow{2}{*}{$\texttt{EnhanceNet}$} & \multirow{2}{*}{$\texttt{AGCRN}$} & \multirow{2}{*}{$\texttt{ST-WA}$}  \\ 
\cline{1-2}
Dataset                                                                          & Metric &                                    &                                        &                                   &                                    \\ 
\hline
\multirow{3}{*}{\begin{tabular}[c]{@{}c@{}}\textbf{PEMS03} \\N=358\end{tabular}}                                         & MAE    & 24.08                              & 23.42                                  & 22.17                             & \textbf{20.87}                     \\
                                                                                 & MAPE   & 23.89                              & 23.69                                  & 23.11                             & \textbf{22.33}                     \\
                                                                                 & RMSE   & 39.91                              & 38.99                                  & 37.10                             & \textbf{35.53}                     \\ 
\hline
\multirow{3}{*}{\begin{tabular}[c]{@{}c@{}}\textbf{PEMS04} \\N=307\end{tabular}} & MAE    & 25.79                              & 28.72                                  & 57.41                             & \textbf{23.54}                     \\
                                                                                 & MAPE   & 18.67                              & 20.68                                  & 47.40                             & \textbf{16.52}                     \\
                                                                                 & RMSE   & 39.87                              & 45.20                                  & 86.02                             & \textbf{36.87}                     \\ 
\hline
\multirow{3}{*}{\begin{tabular}[c]{@{}c@{}}\textbf{PEMS07} \\N=883\end{tabular}}                                         & MAE    & \multirow{3}{*}{OOM}               & \multirow{3}{*}{OOM}                   & 49.64                             & \textbf{25.04}                     \\
                                                                                 & MAPE   &                                    &                                        & 23.43                             & \textbf{10.85}                     \\
                                                                                 & RMSE   &                                    &                                        & 73.50                             & \textbf{38.62}                     \\ 
\hline
\multirow{3}{*}{\begin{tabular}[c]{@{}c@{}}\textbf{PEMS08} \\N=170\end{tabular}}                                       & MAE    & 22.74                              & 37.56                                  & 46.50                             & \textbf{19.27}                     \\
                                                                                 & MAPE   & 15.59                              & 23.31                                  & 19.05                             & \textbf{13.30}                     \\
                                                                                 & RMSE   & 35.95                              & 64.51                                  & 74.78                             & \textbf{30.69}                     \\
\bottomrule
\end{tabular}


       \vspace{5pt}
      \caption{Overall Accuracy, $H=72$, $U=72$.}
      \label{table:results_72}
    \end{minipage}%
    \begin{minipage}{.5\linewidth}
      \centering
\setlength\tabcolsep{2pt}
\centering
\begin{tabular}{|c|c|ccc|ccc|} 
\toprule
\multicolumn{2}{|c|}{Baseline methods}    & \multirow{2}{*}{$\texttt{GRU}$} & \multirow{2}{*}{$\texttt{GRU+S}$} & \multirow{2}{*}{$\texttt{GRU+ST}$} & \multicolumn{1}{l}{\multirow{2}{*}{$\texttt{ATT}$}} & \multirow{2}{*}{$\texttt{ATT+S}$} & \multirow{2}{*}{$\texttt{ATT+ST}$}  \\ 
\cline{1-2}
Dataset                          & Metric &                                 &                                   &                                    & \multicolumn{1}{l}{}                                &                                   &                                     \\ 
\hline
\multirow{3}{*}{\textbf{PEMS03}} & MAE    & 19.97                           & 18.73                             & \textbf{18.68 }                    & 20.03                                               & 18.08                             & \textbf{17.87 }                     \\
                                 & MAPE   & 19.62                           & 20.67                             & \textbf{17.61 }                    & 20.01                                               & 17.10                             & \textbf{17.00 }                     \\
                                 & RMSE   & 32.77                           & 31.22                             & \textbf{31.08 }                    & 32.71                                               & 30.15                             & \textbf{29.86 }                     \\ 
\hline
\multirow{3}{*}{\textbf{PEMS04}} & MAE    & 26.02                           & 24.48                             & \textbf{23.95 }                    & 25.84                                               & 23.33                             & \textbf{23.09 }                     \\
                                 & MAPE   & 17.23                           & 17.11                             & \textbf{15.59 }                    & 17.18                                               & 16.13                             & \textbf{15.57 }                     \\
                                 & RMSE   & 40.11                           & 37.68                             & \textbf{37.18 }                    & 39.99                                               & 36.63                             & \textbf{36.61 }                     \\ 
\hline
\multirow{3}{*}{\textbf{PEMS07}} & MAE    & 27.60                           & 26.08                             & \textbf{25.93 }                    & 27.39                                               & 24.57                             & \textbf{24.47 }                     \\
                                 & MAPE   & 12.41                           & 11.55                             & \textbf{11.34 }                    & 12.05                                               & 10.99                             & \textbf{10.72 }                     \\
                                 & RMSE   & 43.28                           & 40.70                             & \textbf{40.68 }                    & 43.41                                               & 39.58                             & \textbf{39.55 }                     \\ 
\hline
\multirow{3}{*}{\textbf{PEMS08}} & MAE    & 20.75                           & 19.19                             & \textbf{19.02 }                    & 20.54                                               & 17.95                             & \textbf{17.76 }                     \\
                                 & MAPE   & 13.20                           & 12.44                             & \textbf{12.40 }                    & 14.11                                               & 11.92                             & \textbf{11.85 }                     \\
                                 & RMSE   & 32.15                           & 29.75                             & \textbf{29.41 }                    & 32.01                                               & 28.45                             & \textbf{28.30 }                     \\
\bottomrule
\end{tabular}
         \vspace{5pt}
         \captionsetup{justification=centering}
\caption{Overall Accuracy of Enhanced versions of \texttt{RNN} and \texttt{ATT}, $H=12$, $U=12$}
        \label{table:results_temporal}
    \end{minipage} 
    \vspace{-15pt}
\end{table*}

\subsection{Experimental Results}\label{section:results}

\noindent
\textbf{Overall Accuracy.} Table~\ref{table:results} shows overall accuracy on the default forecasting setup in which $H=12$ and $U=12$. First, we observe that temporal-aware model \texttt{meta-LSTM} performs the worst. This is due to that it is the only baseline which does not explicitly model the sensor correlations. \revision{Next, we observe that the spatial-aware methods, i.e. $\texttt{EnhanceNet}$ and $\texttt{AGCRN}$, outperform most of the ST-agnostic baselines, confirming our analysis that using a shared parameter space for time series from different locations is not optimal when modeling spatial-temporal dynamics. 
Furthermore our proposed method $\texttt{ST-WA}$ achieves the best accuracy on 10 out of 12 metrics over four data sets, which justifies our design choices and highlights the importance of explicitly modeling spatio-temporal model parameters.} 

Next, to study the ability of forecasting long into the future we move away from the classic 1h to 1h setting and increase both $H$ and $U$ to 72 which represents 6 hours. The results are shown in Table \ref{table:results_72}. We only compare \texttt{ST-WA} with $\texttt{AGCRN}$, $\texttt{EnhanceNet}$ and $\texttt{STFGNN}$, which are the top-3 baselines from the previous experiment. We can observe that our proposed method significantly outperforms the baselines.  Furthermore we observe that $\texttt{EnhanceNet}$ and $\texttt{STFGNN}$ run out of memory (shown as OOM in Table \ref{table:results_72}) 
when the number of sensors $N$ is big, e.g., 883 in the case of $\textbf{PEMS07}$. In contrast, \texttt{ST-WA} still works well, suggesting its high memory efficiency.

\noindent
\textbf{Model-Agnostic ST-Aware Model Parameter Generation. }
In this experiment, we demonstrate that the proposed ST-Aware Model Parameter Generation is model-agnostic, which is able to be applied to different types of forecasting models and consistently enhance their accuracy. We consider two types of forecasting models---\texttt{GRU} and self-attention based Transformer \texttt{ATT}. We first compare the base \texttt{GRU} and \texttt{ATT} with their enhanced versions $\texttt{+S}$ denotes \textit{spatial-aware} and  $\texttt{+ST}$ stands for \textit{spatial-temporal aware} variants.
Table \ref{table:results_temporal} shows the accuracy. We can see that $\texttt{+S}$ offers clear improvements and $\texttt{+ST}$ further improves. This proves that our proposal is model-agnostic and it can easily be integrated into existing methods to further improve their accuracy.

\noindent
\textbf{Ablation Study.} We perform an ablation study on the {PEMS04} dataset by removing different components from the full model $\texttt{ST-WA}$. First, $\texttt{S-WA}$ is a model by removing the $\textbf{z}^{(i)}_{t}$ from the full model. This corresponds to a model where weight generation only generates location-specific but not time-varying  model parameters. Second, $\texttt{WA}$ is stacked Window Attention without ST-aware projection matrix generation.  
Third, $\texttt{WA-1}$ is Window Attention, by removing the stacked structures from \texttt{WA}, i.e., a single layer of window attention. Third, we consider $\texttt{SA}$ that employs canonical self-attention, i.e., Transformers.  
\begin{table}[h!]
\centering

\begin{tabular}{|c|rrrrr|} 
\toprule
                               & $\texttt{SA}$  & $\texttt{WA-1}$ & $\texttt{WA}$ & \multicolumn{1}{c}{$\texttt{S-WA}$} & $\texttt{ST-WA}$  \\ 
\hline
MAE                            & 23.31          & 19.56           & 19.38         & 19.25                               & \textbf{19.06}   \\
MAPE                           & 15.69          & 12.96           & 12.89         & 12.81                               & \textbf{12.52}   \\
RMSE                           & 36.76          & 31.55           & 31.29         & 31.37                               & \textbf{31.02}   \\
Memory                         & 23.94          & \textbf{3.03}  & 4.44          & 8.02                                & 8.14              \\
\multicolumn{1}{|l|}{Training} & 44.22          & \textbf{7.88}            & 13.25         & 21.62                               & 21.93             \\
\# Para                        & 393k & \textbf{164k}            & 490k          & 392k                                & 396k              \\
\bottomrule
\end{tabular}

\caption{Ablation study on \textbf{PEMS04}} 
\label{table:ablation}
\vspace{-4pt}
\end{table}

\begin{figure}[H]
    \vspace{-5pt}
     \centering
     \includegraphics[width=0.55\linewidth]{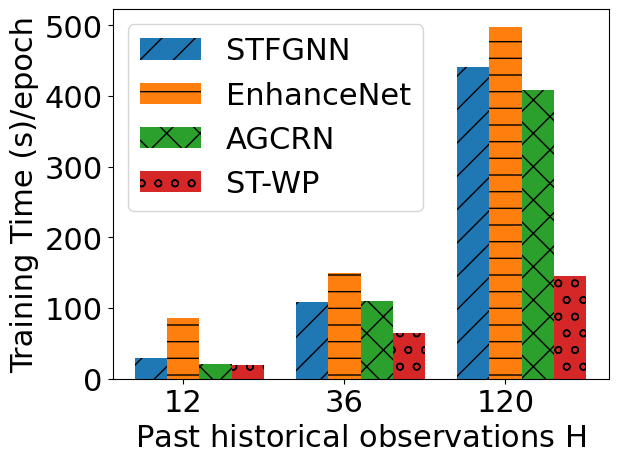}
     \caption{Runtime, \textbf{PEMS04}}
     \label{fig:runtime}
     \vspace{-5pt}
\end{figure}

Table \ref{table:ablation} shows the accuracy along with training time (seconds per epoch), memory usage (GB), and number of parameters to be learned. 
First, when comparing $\texttt{SA}$ with $\texttt{WA-1}$, $\texttt{WA-1}$ is more than 3x faster and requires 5x less memory, suggesting the window attention is accurate and efficient, compared to canonical self-attention. $\texttt{WA}$ further improves accuracy over $\texttt{WA-1}$, suggesting that the multi-layer hierarchical structure is effective. $\texttt{S-WA}$ and $\texttt{ST-WA}$ achieve a clear accuracy improvement, and $\texttt{ST-WA}$ achieves the highest accuracy, highlighting the need of having both spatial and temporal aware model parameters. For run-time, both models take longer time, while sill being comparable with other baselines (c.f. Figure \ref{fig:runtime}). We conclude that, when training run-time is critical, we recommend the use of $\texttt{WA-1}$ or $\texttt{WA}$. Otherwise, $\texttt{ST-WA}$ is the best choice. 
 
 \begin{figure*}
\hfill
    \begin{subfigure}[t]{0.34\textwidth}
         \centering
         \includegraphics[height=3.5cm]{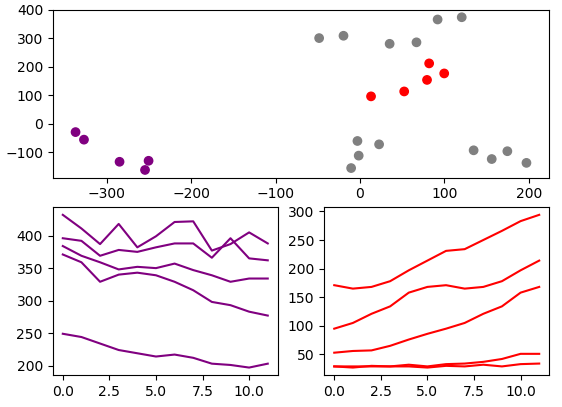}
          \caption{2D Points of $\phi^{(i)}_t$ vs. Input Time Series $\textbf{x}^{(i)}_t$}
        \label{fig:patterns}
    \end{subfigure}
    \begin{subfigure}[t]{0.28\textwidth}
         \fbox{\includegraphics[height=3.3cm]{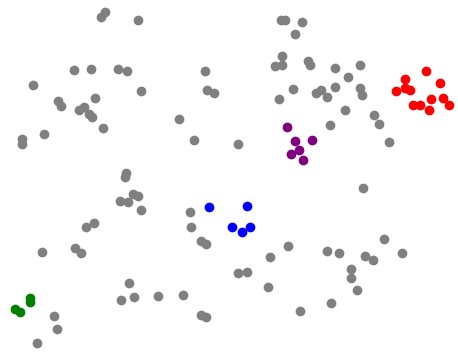}}
          \caption{2D Points of $\mathbf{z}^{(i)}$}
        \label{fig:tsne}
     \end{subfigure}
    \begin{subfigure}[t]{0.35\textwidth}
         \centering
         \includegraphics[height=3.5cm]{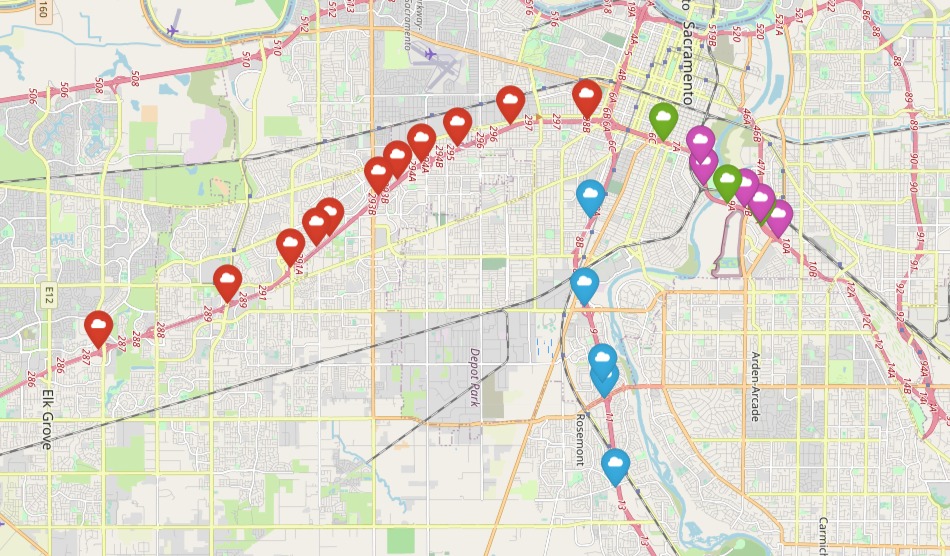}
         \caption{Sensor Locations.}
        \label{fig:map}
     \end{subfigure}
     \caption{Justifying Design Choices with Visualization. } 
     \vspace{-10pt}
\end{figure*}

\noindent
\textbf{Impact of Long Historical Window $H$:} 
We systematically increase $H$ from 12 to 36 and 120. The results are shown in Table  \ref{table:long_range_res}. \texttt{ST-WA} shows improvements with longer $H$ while the other baselines does not improve significantly, some even loosing accuracy. This suggests that the proposed model is able to better capture long term dependencies while other baselines struggle to do so.

\noindent
\textbf{Impact of Window Size $S$: } To investigate the effect of the window size $S$, we run multiple experiments in which we change the window size and the total number of layers of $\texttt{ST-WA}$. The results are shown in Table \ref{table:patches}. We observe that, when using the window size equal to the length of the sequence $S=H=12$, it performs the worst. This is understandable, as it only uses 1 layer, the total number of parameters is significantly smaller resulting in lower representation power. Next, we can see that when using 3 layers, there are small variations among different window size settings, suggesting the proposed method is insensitive to window sizes.

\begin{table}[h]
\centering

\begin{tabular}{|l|c|c|c|c|l|l|} 
\toprule
\multirow{2}{*}{} & \multicolumn{3}{c|}{3 layers}                                                              & \multicolumn{2}{c|}{2 layers}      & 1 layer  \\ 
\cline{2-7}
                  & \multicolumn{1}{l|}{S=3,2,2} & \multicolumn{1}{l|}{S=2,3,2} & \multicolumn{1}{l|}{S=2,2,3} & \multicolumn{1}{l|}{S=4,3} & S=6,2 & S=12     \\ 
\hline
MAE               & \textbf{19.06}               & 19.10                        & 19.13                        & 19.62                      & 19.69 & 22.05    \\
MAPE              & \textbf{12.52}               & 12.62                        & 12.58                        & 13.13                      & 12.86 & 14.99    \\
RMSE              & \textbf{31.02}               & 31.05                        & 31.15                        & 31.51                      & 31.81 & 35.35    \\
\bottomrule
\end{tabular}
\caption{Effect of Window Sizes, \textbf{PEMS04}}
\label{table:patches}
\vspace{-3pt}
\end{table}

\noindent
\textbf{Impact of the Regularizer:} To further investigate the effect of the KL based  regularization term, we perform an experiment in which we compare the full model $\texttt{ST-WA}$ with and without using the regularization term in the loss function. We can see from Table \ref{table:reg} that removing the regularization term results in a clear accuracy loss, which justifies our design choice.

\begin{table}[h!]
\centering
\small
\begin{tabular}{|l|c|c|} 
\toprule
 & \multicolumn{1}{l|}{With} & Without  \\ 
\hline
MAE    & \textbf{19.06}            & 19.23    \\
MAPE   & \textbf{12.52}            & 12.62    \\
RMSE   & \textbf{31.02}            & 31.37    \\
\bottomrule
\end{tabular}
\caption{Effect of the Regularization Term,  \textbf{PEMS04}}
\label{table:reg}
\vspace{-3pt}
\end{table}

\noindent
\textbf{Impact of Stochastic Latent Variables:} We perform an additional ablation study in which we compare the proposed method \texttt{ST-WA} with a deterministic version in which we replace the stochastic latent variables, $\textbf{z}^{(i)}$ and $\textbf{z}^{(i)}_t$, with deterministic latent variables, i.e., 16-dimensional vectors. In addition, we remove the KL based regularization from the loss function, as it is no longer needed for deterministic latent variables. We can see from Table \ref{table:results2} that the stochastic version constantly outperforms the deterministic version,  which justifies the effectiveness of our design choice on using stochastic variables.

\begin{table}[h]
\centering
\label{table:long_range}
\begin{tabular}{|c|ccc|} 
\toprule
                                         & MAE                             & MAPE                            & RMSE                              \\ 
\hline
\texttt{ST-WA}                                     &\textbf{19.06}                    &\textbf{12.52}                     &\textbf{31.02}                      \\ 
\hline
\multicolumn{1}{|l|}{Deterministic \texttt{ST-WA}} & 19.32 & 12.72 & 31.41  \\
\bottomrule
\end{tabular}
\caption{Effect of stochastic latent variables,  \textbf{PEMS04}}

\label{table:results2}
\vspace{-3pt}
\end{table}

\textbf{Impact of Stochastic Latent Variables Size $k$:} 
We study the impact of the size of stochastic latent variable, i.e., $k$. 
Recall that the stochastic latent variables follow a multivariate Gaussian distribution in a $k$-dimensional space. We perform an additional experiment in which we vary $k$, i.e., the size of $\textbf{z}^{(i)}$ and $\textbf{z}^{(i)}_t$ among $\{4,8,16,32\}$. The results are shown in Table \ref{table:memory_size}. We can observe that when the latent variable size is small, i.e. 4, there is a significant drop in performance, as 4-dimensional Gaussian distributions might not be sufficient to fully capture the dynamics of the traffic conditions. When the size is too large, i.e., 32, we see again a big drop in performance which is due to 
over-fitting. 




\begin{table}[h]
\centering
\begin{tabular}{|c|ccc|} 
\toprule
$k$ & MAE                                       & MAPE                                      & RMSE                     \\ 
\cmidrule{1-2}\cline{3-4}
4   & 19.42                                     & 12.62                                     & 31.02                    \\
8   & 19.37                                     & 12.58                                     & 31.15                    \\
16  & \textbf{\textbf{\textbf{\textbf{19.06}}}} & \textbf{\textbf{\textbf{\textbf{12.52}}}} & \textbf{\textbf{31.02}}  \\
32  & 19.42                                     & 12.96                                     & 31.09                    \\
\bottomrule
\end{tabular}

\caption{Effect of Latent Stochastic Variables Size $k$, \textbf{PEMS04}}
\label{table:memory_size}
\end{table}
\vspace{-5pt}

\noindent
\textbf{Impact of Number of Proxies $p$:} To study the impact of number of proxies $p$ we have conducted an additional ablation study in which we vary $p$ from 1, 2, to 3. The results are shown in Table \ref{table:proxy}. We can observe that the bigger the number of proxies is, the better the accuracy is, but this accuracy improvement comes by compromising on training efficiency (s/epoch) and memory usage on the parameters.

\begin{table}[h!]
\centering

\begin{tabular}{|c|ccccc|} 
\toprule
$p$ & MAE   & MAPE  & RMSE  & Training & \# Para  \\ 
\hline
1     & 23.93 & 16.48 & 37.44 & \textbf{69.15}    & \textbf{606k}     \\
2     & 23.54 & 16.52 & 36.87 & 102.93   & 764k     \\
3     & \textbf{23.48} & \textbf{16.50} & \textbf{36.70} & 128.25   & 922k     \\
\bottomrule
\end{tabular}
\caption{Effect of number of Proxies, \textbf{PEMS04}}
\label{table:proxy}
\vspace{-6pt}
\end{table}
\noindent
\textbf{Impact of Aggregation Function:} To study the impact of the aggregator function, we ran an additional experiment in which we compare with a mean aggregator, where each proxy within a window is weighted equally. We can observe from Table \ref{table:aggregator} that when using a simple mean aggregator the accuracy drops significantly, which justify our design choice (cf. Equation~ \ref{eq:weight_network}).

\begin{table}[h!]
\centering
\begin{tabular}{|c|ccc|} 
\toprule
                & MAE   & MAPE  & RMSE   \\ 
\hline
Mean Aggregator & 24.65 & 17.21 & 38.41  \\
Our Aggregator  & \textbf{23.54} & \textbf{16.52} & \textbf{36.87}  \\
\bottomrule
\end{tabular}

\captionsetup{justification=centering}
\caption{Effect of different Aggregation Functions, \textbf{{PEMS04}}}
\label{table:aggregator}
\vspace{-3pt}
\end{table}

\noindent
\textbf{Run-time:} To study the efficiency of our proposed method we have reported the training run-time (seconds/epoch) by systematically increasing the historical window $H$ from 12 to 36 and 120, corresponding to 1, 3, and 12 hours respectively. The results are shown in Figure \ref{fig:runtime}. We can observe that for short  historical windows $\texttt{STFGNN}$, $\texttt{AGCRN}$ and $\texttt{ST-WA}$ have similar run-time, while $\texttt{EnhanceNet}$ takes significantly longer to train. In addition we can see an exponential growth in run-time w.r.t the sequence length for all baselines. In contrast $\texttt{ST-WA}$ is significantly more efficient which increases its usability to applications where processing long-sequence data is required.


\subsection{Visualization of Learned Stochastic Variables}

To investigate if the learned stochastic variables are able to capture different patterns during different time windows, we use \texttt{t-SNE}~\cite{tsne} to embed the projection matrices $\phi^{(i)}_t=\{\mathbf{Q}{t}^{(i)}, \mathbf{K}_{t}^{(i)}, \mathbf{V}_{t}^{(i)}\}$ during different time windows from a specific time series to 2D points, because the projection matrices are generated from the learned stochastic variables. From Figure \ref{fig:patterns}, we observe that the 2D points spread over the space, indicating that different projection matrices are utilized to capture different patterns during different time windows. Next, we have selected two clusters of the 2D points and show their input time series in the corresponding windows. We observe that the red and purple point clusters specialise in identifying up/down trends, respectively. This provides evidence that the proposal is able to capture well temporal dynamics.

We proceed to investigate if the learned location-specific stochastic latent variable $\textbf{z}^{(i)}$ capture distinct, and the most prominent patterns of time series from different locations. We use \texttt{t-SNE}~\cite{tsne} to compress each time series's $\textbf{z}^{(i)}$ to a 2D space, which is show in Figure~\ref{fig:tsne}. We observe that the points are spread over the space, indicating that time series from different locations have their unique patterns. We use four different colors to highlight four clusters of the 2D points. Then, we show their  physical locations of the deployed sensors on a map in Figure~\ref{fig:map}. We observe that each cluster contains sensors deployed along the same road. This is intuitive since traffic patterns are expected to follow similar patterns along the same street. In addition, we observe that green and purple sensors are close in Figure \ref{fig:map} but their 2D points are far in Figure \ref{fig:tsne}. This is because the sensors in the purple cluster are deployed on the road that goes from south to north, where the sensors in the green cluster are on the opposite direction. Two different directions on the same road may have quite different patterns. These provide strong evidence that the learned stochastic variables well reflect the patterns of time series from different locations.





\section{Conclusion and Outlook}\label{section:conclusion}

We propose a data-driven, model-agnostic method to turn spatio-temporal agnostic models to spatio-temporal aware models. In particular, we apply the method to enable spatio-temporal aware attention that is capable of modeling complex spatio-temporal dynamics in traffic time series. In addition, we propose an efficient window attention with linear complexity, thus ensuring competitive overall efficiency. Extensive experiments on four datasets show that our proposal outperforms other state-of-the-art methods. A limitation of our proposal is that the learning is based on the assumption that the latent stochastic variables follow Gaussian distributions.  In future research, it is of interest to explore methods such as normalizing flows for to employ non-Gaussian stochastic variables. It is also of interest to explore the parallelism of the proposal~\cite{DBLP:conf/waim/YuanSWYZY10}.   

\section*{Acknowledgements}
This work was supported in part by Independent Research Fund Denmark under agreements 8022-00246B and 8048-00038B, the VILLUM FONDEN under agreements 34328 and 40567, Huawei Cloud Database Innovation Lab, and the Innovation Fund Denmark centre, DIREC.

\bibliographystyle{IEEEtran}
\bibliography{references}

\end{document}